\documentclass{article}
\usepackage{amsmath}

\usepackage{arxiv}

\usepackage[utf8]{inputenc} 
\usepackage[T1]{fontenc}    
\usepackage{hyperref}       
\usepackage{url}            
\usepackage{booktabs}       
\usepackage{amsfonts}       
\usepackage{nicefrac}       
\usepackage{microtype}      
\usepackage{cleveref}       
\usepackage{lipsum}         
\usepackage{graphicx}
\usepackage{doi}
\usepackage{subcaption}
\usepackage{ragged2e}

\title{Robust Physics-Informed Neural Network Approach for Estimating Heterogeneous Elastic Properties from Noisy Displacement Data

}
\usepackage{authblk}

\newbox{\orcid}\sbox{\orcid}{\includegraphics[scale=0.06]{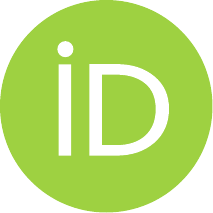}} 
\author[1]{%
	{\hspace{1mm}Tatthapong~Srikitrungruang}%
}
\author[1]{%
	{\hspace{1mm}Matthew~Lemon}%
}
\author[1]{%
	{\hspace{1mm}Sina~Aghaee~Dabaghan~Fard}%
}
\author[1]{%
	\href{https://orcid.org/0000-0001-7533-1865}{\usebox{\orcid}\hspace{1mm}Jaesung~Lee\thanks{\texttt{j.lee@tamu.edu}}}%
}
\author[2]{%
	{\hspace{1mm}Yuxiao Zhou}%
}
\affil[1]{Wm Michael Barnes ’64 Department of Industrial and Systems Engineering, Texas A\&M University, College Station, TX, 77843, USA}
\affil[2]{Mike Walker ’66 Department of Mechanical Engineering, Texas A\&M University, College Station, TX, 77840, USA}
\renewcommand{\shorttitle}{Inverse Elasticity Physics-informed Neural Network (IE-PINN)}
\hypersetup{
pdftitle={Robust Physics-Informed Neural Network Approach for Estimating Heterogeneous Elastic Properties from Noisy Displacement Data

},
pdfsubject={},
pdfauthor={Tatthapong~Srikitrungruang, Jaesung~Lee},
}

\date{}

\usepackage{bm}
\newcommand{\boundarySpace}{S}

\begin{document}

\title{Robust Physics-Informed Neural Network Approach for Estimating Heterogeneous Elastic Properties from Noisy Displacement Data

}

\maketitle
\begin{abstract}

     \large Accurately estimating spatially heterogeneous elasticity parameters, particularly Young's modulus and Poisson’s ratio, from noisy displacement measurements remains significantly challenging in inverse elasticity problems. Existing inverse estimation techniques are often limited by instability, pronounced sensitivity to measurement noise, and difficulty in recovering absolute-scale Young's modulus. This work presents a novel Inverse Elasticity Physics-Informed Neural Network (IE-PINN) specifically designed to robustly reconstruct heterogeneous distributions of elasticity parameters from noisy displacement data based on linear elasticity physics. IE-PINN integrates three distinct neural network architectures dedicated to separately modeling displacement fields, strain fields, and elasticity distributions, thereby significantly enhancing stability and accuracy against measurement noise. Additionally, a two-phase estimation strategy is introduced: the first phase recovers relative spatial distributions of Young’s modulus and Poisson’s ratio, and the second phase calibrates the absolute scale of Young’s modulus using imposed loading boundary conditions. Additional methodological innovations, including positional encoding, sine activation functions, and a sequential pretraining protocol, further enhance the model’s performance and robustness. Extensive numerical experiments demonstrate that IE-PINN effectively overcomes critical limitations encountered by existing methods, delivering accurate absolute-scale elasticity estimations even under severe noise conditions. This advancement holds substantial potential for clinical imaging diagnostics and mechanical characterization, where measurements typically encounter substantial noise.

\end{abstract}


\setcounter{page}{1}
\section{Introduction}
Quantifying the spatial distributions of elastic properties, specifically Young's modulus and Poisson's ratio is crucial in numerous applications, including biomedical imaging. 
Young's modulus characterizes a material's local resistance to elastic (reversible) axial deformation, while Poisson's ratio quantifies the coupling between axial and transverse deformations. 
When external loads are applied, an internal stress field is established within the solid material, resulting in deformation (i.e., displacement) patterns that depend directly on spatial distributions of the elastic properties. 
Accurate spatial characterization of elastic properties enables precise disease diagnosis (e.g., cardiovascular and airway diseases,\cite{boutouyrie2021arterial, chirinos2019large, wang2024sensory, kim2024mucosa, han2023actuation} cancer.\cite{Goenezen2012, Quan2016, Wu2019}), development biomedical devices and tissue engineering scaffolds,\cite{wang2023situ, xue2022biodegradable, zhou2022three} assessment of structural integrity (e.g., bone health,\cite{wang2013trabecular} engineered and additive manufactured parts.\cite{Choren2013, Roy2020, Yan2024}), and precise computational mechanics modeling.\cite{mao2019voxel, su2021effects}
Typically, elastic properties vary spatially within these materials, necessitating advanced methods to characterize their heterogeneity.

Techniques for estimating elastic properties are generally classified into direct and indirect methods.  
Direct methods, such as nanoindentation and atomic force microscopy, quantify elasticity distributions by generating local deformation on the sample surface with a known force.\cite{Oliver1992, Oliver2004, Xu2022,zhou2022atomic} 
However, these direct methods are limited to local areas on exposed surfaces, often requiring destructive procedures to access areas of interest, making them impractical for many applications.
Indirect methods infer the elasticity distributions by measuring the displacement field resulting from externally applied force to the surface. 
Medical imaging techniques, such as Magnetic Resonance Elastography (MRE),\cite{muthupillai1995magnetic, manduca2021mr} and ultrasound-based elastography methods,\cite{Lang1970, Buntin1990} are commonly employed indirect methods.
Additionally, Digital Image Correlation (DIC) and Digital Volume Correlation (DVC) techniques are employed during mechanical testing to extract displacement fields from 2D and 3D imaging data, respectively.\cite{MichaelA2009, bay1999digital, chen2024engineering, tanoto2025predicting, tanoto2024quantifying} 
Materials with spatially varying elasticity often show complex patterns in displacement measurements and significant noise.\cite{Gao2003, paraskevoulakos2024sensitivity} 
Conventionally, strain-based elastography methods have been widely used in clinical applications. By assuming a uniform stress distribution, they obtain elasticity distributions in relative scale by inverting strains. To achieve uniform stress, trained technicians apply load uniformly over the surface. However, even uniform loading leads to heterogeneous stress distributions within the material, which is known as stress localization.

To achieve more robust and precise elasticity property estimation from the displacement field, physics-informed approaches incorporate governing physical principles typically formulated as partial differential equations (PDEs) models, such as those describing linear elasticity.\cite{Gould2013}
Estimating elasticity using PDE models is inherently ill-posed, often resulting in unstable or non-unique solutions.\cite{Kabanikhin2008}
These approaches are categorized into direct and iterative methods. 
Direct methods simplify or reformulate linear elasticity PDE equations into forms solvable analytically or numerically, typically limited to simple geometries or idealized conditions.\cite{Barbone2007, Barbone2009, Babaniyi2017} 
These methods generally require smooth displacements and strain fields without noise and auxiliary constraints such as average Young's modulus. Minor errors from noise in the data or inaccuracies in the constraints can propagate through estimation, degrading the accuracy and stability of the solutions.\cite{Barbone2007}
The iterative methods utilize finite element models (FEM) to minimize discrepancies between measured and simulated outputs iteratively.\cite{Doyley2000, Oberai2003, Oberai2004, Doyley2005, Smyl2019} The iterative methods are computationally intensive and heavily dependent on the given initial conditions. 
In practice, displacement measurements often contain noise, from which direct strain calculation with numerical differentiation amplifies the errors, hindering accurate elasticity estimation.\cite{Sutton2008, Pan2013, Freddi2015}
Additionally, many methods typically assume incompressibility, which is impractical for many solid materials that exhibit compressibility and heterogeneous Poisson's ratios.\cite{Chen2021,Haghighat2021,Gao2022}

Recent advancements in machine learning (ML) methods, 
such as Gaussian process regression (GPR),\cite{Nguyen2022} deep neural networks (generative adversarial networks,\cite{Ni2021}  and convolutional neural networks.\cite{mouloodi2020prediction}) have enabled elasticity mapping from direct elasticity measurements.  
However, these purely data-driven methods typically exhibit limited generalization and reduced accuracy when applied to materials with complex heterogeneity or geometries.

Physics-informed neural networks (PINN), integrating physical laws directly into neural network architectures, have demonstrated success primarily in forward elasticity problems, predicting displacement or strain with given known parameters.\cite{Raissi2019, Raissi2020, Song2022}
Nonetheless, inverse elasticity estimation using PINN remains significantly more challenging. 
Most existing inverse PINN approaches assume homogeneous elasticity with a constant value to be estimated.\cite{Haghighat2021, Gao2022, Sahin2024, Mallampati2021, Min2024,  Kianian2024, Hamel2022}
Heterogeneous elasticity estimation is generally challenging as, at every point, different elasticity values need to be estimated. 
Recent works addressing heterogeneous elasticity estimation with PINNs  frequently struggle with noisy displacement data and often produce relative rather than absolute Young's modulus values.\cite{Chen2021,Shukla2022, Ragoza2023, Kamali2023, Kamali2024,  Chen2023} To simplify the problem, strict assumptions are often made, such as known true (internal or boundary) stress distributions,\cite{Kamali2023, Kamali2024} known mean Young's modulus,\cite{Chen2021, Chen2023} and incompressible materials.\cite{Chen2021,Ragoza2023}

In this paper, we propose an Inverse Elasticity PINN (IE-PINN) model specifically designed to estimate the spatial distributions of elasticity parameters, namely, Young's modulus and Poisson's ratio, from noisy displacement data based on the governing physics of linear elasticity. Our IE-PINN model demonstrates robust performance against noise, enabling accurate recovery of absolute Young's modulus distribution rather than merely relative distribution. 
A novel two-step approach incorporating loading force boundary conditions facilitates precise absolute elasticity estimation. 
By combining with a specialized neural network architecture, our method achieves robust heterogeneous elasticity estimation with low errors based on noisy displacement datasets.

\section{Result and Discussion}
\label{sec:Result}

\subsection{Inverse Elasticity Physic-informed Neural Network (IE-PINN)}

Estimating Young’s modulus and Poisson’s ratio from deformation fields constitutes an inherently ill-posed inverse elasticity problem, often yielding non-unique, unstable, and noise-sensitive solutions. 
Conventional PINNs mainly target forward problems, benefiting from the automatic differentiation needed for PDE. However, automatic differentiation is an ineffective inverse elasticity problem (Figure \ref{Supple_FDADLoss} in Supporting Information). 
Recently, Elastnet successfully estimates heterogeneous elasticity based on ideal noise-free displacements;\cite{Chen2023} however, it directly applies finite differentiation to displacement measurement data without functional approximation, making it vulnerable and unstable to even small noise (Section \ref{ssec:noise sensitivity}). Additionally, Elastnet estimates only relative Young's modulus distributions,\cite{Chen2021, Chen2023} requiring the true mean Young's modulus to derive absolute values. However, the true mean Young's modulus is typically not available in practice.

\begin{figure}[t!]
    \centering
    \includegraphics[width=\textwidth]{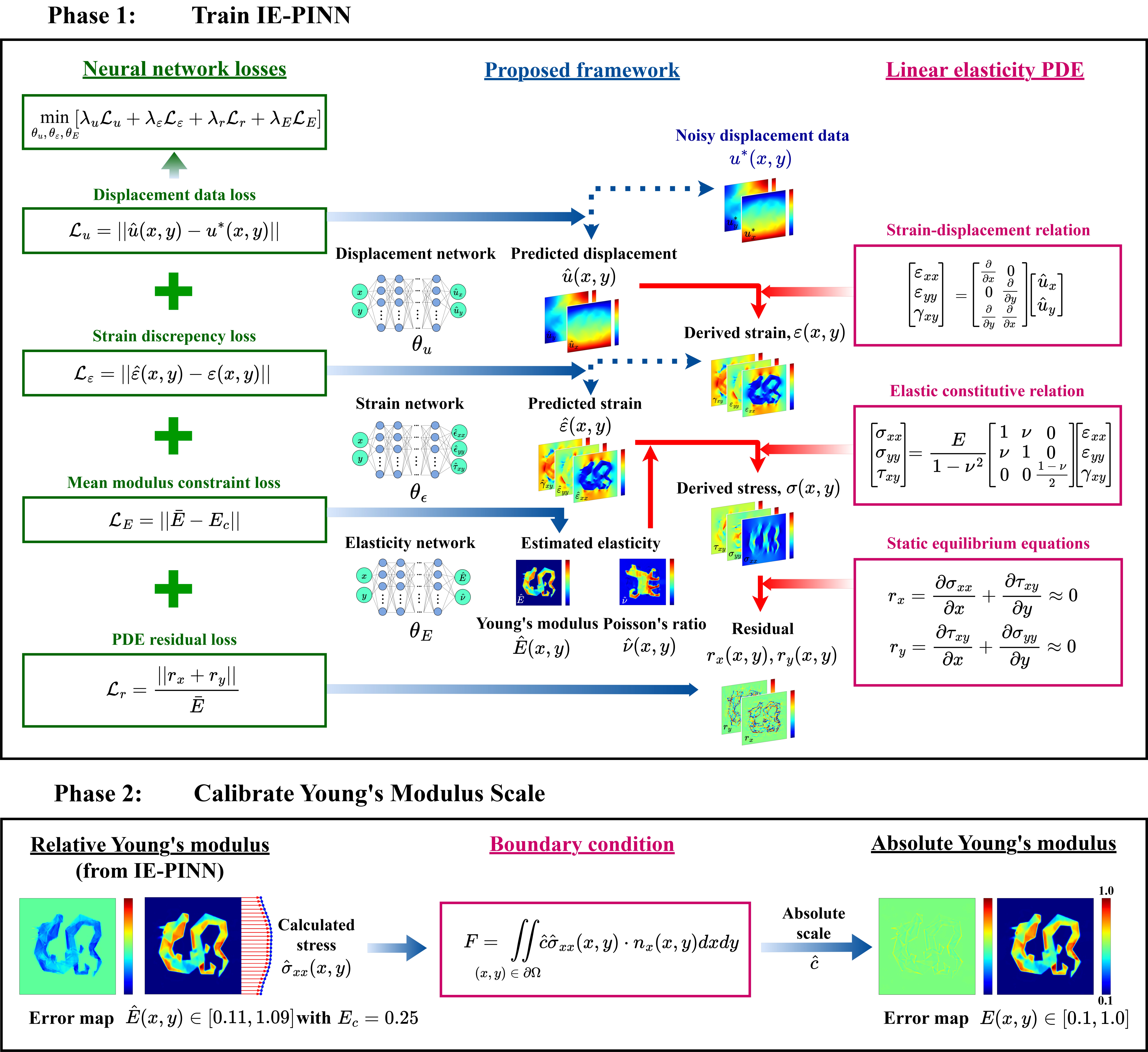}
    \vspace{0.2 in}
    \justifying
    \caption{Framework for heterogeneous elasticity estimation from noisy displacement data. The framework consists of two distinct phases: the Inverse Elasticity Physics-informed Neural Network (IE-PINN) training phase and the Young's modulus scale calibration phase. In the first phase, three neural networks are trained using spatial coordinates to predict mechanical quantities such as displacement, strain, and elasticity (Young's modulus and Poisson's ratio), respectively.
    The displacement network is specifically employed to mitigate the adverse impact of noisy displacement measurements. The predicted displacements are used to compute the strain vector via the strain-displacement relation. The strain network fits the strain derived by the displacement network. The elasticity network predicts Young's modulus and Poisson's ratio from spatial coordinates.  Based on the constitutive equation, the stress tensor is derived from the strain tensor, Young's modulus, and Poisson's ratio. The static equilibrium loss is evaluated by finite differentiation of the stress field. 
    The goal of the training process is to estimate the parameters (\(\theta_u,\theta_{\varepsilon},\theta_E\)) of all three neural networks by minimizing the (total) neural network loss, which evaluates the displacement network fitting, discrepancies in strain, deviations in mean modulus constraints, and the partial differential equation (PDE) residuals related to equilibrium equations. In Phase 2, the relative stress predicted at the boundary from Phase 1, combined with the experimental loading boundary conditions, is used to recover the correct absolute scale $\hat{c}$, resulting in an absolute-scale distribution of Young's modulus.}
\label{fig:2_1_1_model_overview}
\end{figure}

To overcome the critical limitations of noise sensitivity and the inability to estimate absolute Young's modulus scales, we propose the IE-PINN framework for robustly estimating heterogeneous Young's modulus and Poisson's ratio distributions from noisy displacement fields. Figure \ref{fig:2_1_1_model_overview} illustrates the proposed methods, consisting of two phases. In Phase 1, IE-PINN is trained based on noisy displacement data, and the spatial distributions of relative Young's modulus and Poisson's ratio are estimated. Then, in Phase 2, the absolute scale of Young's modulus is estimated (referred to as calibration), recovering the absolute scales of Young's modulus distribution by leveraging the predicted relative stress distributions from IE-PINN.

To clearly demonstrate the advantages of IE-PINN, throughout Section \ref{sec:Result}, we employ a dataset from Elastnet study,\cite{Chen2023} where the true spatial distribution of Young's modulus adopts a dragon shape and Poisson's ratio a dog shape, with displacements simulated using FEM. Specifically, this study uses noisy displacement data generated by adding zero-mean Gaussian noise to the displacements. The standard deviation of the noise is set to 0.1\% of the average displacements (i.e., signal-to-noise, or SNR, ratio of 1000 as described in Supplementary Note S1 in Supporting Information) to illustrate performance under mild noise conditions.

\begin{figure}[t!]
    \centering
    \includegraphics[width=\textwidth]{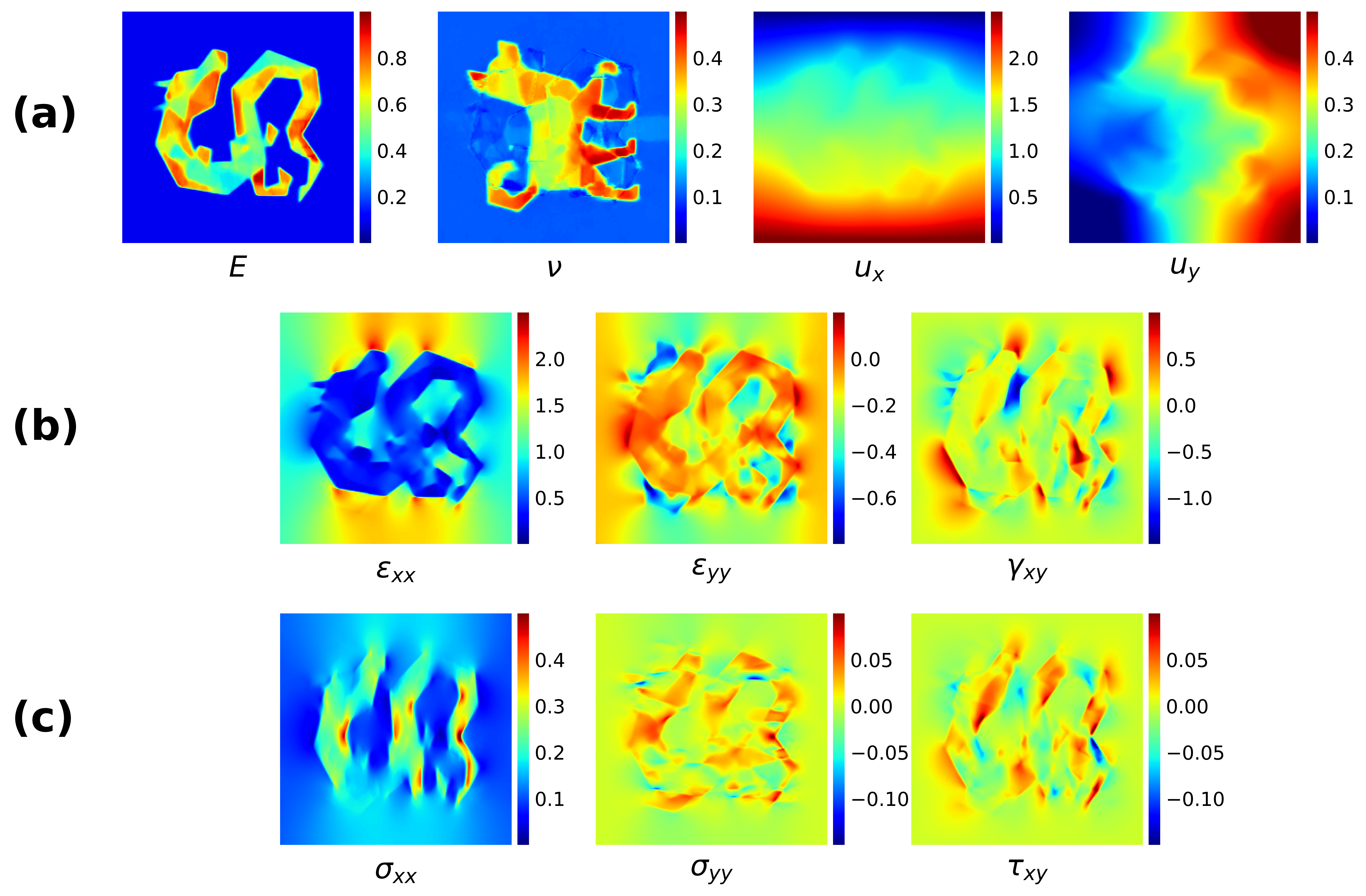}
    \justifying
    \caption{The prediction field of mechanical quantities. The model is applied to a measured displacement that contains a signal-to-noise ratio of 1000. (a) The predicted Young's modulus field (MPa), Poisson's ratio field, and axial displacement field (mm). (b) The predicted strain field ($\%$). (c) The predicted stress field (MPa).}
    \label{fig: 2_1_2_dragon_dog_prediction}
\end{figure}

The IE-PINN architecture integrates the governing PDEs of linear elasticity, including strain-displacement relation, elastic constitutive relation, and equilibrium equations. 
The core innovation for achieving robust elasticity estimation in the presence of noisy displacements lies in the neural network framework, which comprises three deep neural networks: the displacement network, the extra strain network, and the elasticity network. 
The displacement network fits noisy displacement data ($u_x$ and $u_y$) with respect to spatial coordinates ($x$ and $y$). 
Neural networks can effectively mitigate the adverse effect of noise through smoothing.\cite{Chen2019} In PINN, such smoothing is regulated by the PDE equations.
Using neural networks fitted to measurements (i.e., displacements) is a common scheme in PINN. However, displacement fitting alone introduces high sensitivity in its second derivatives; fitting errors are amplified and propagated to the second derivative function, making the inverse elasticity estimation unstable, especially when the data are noisy.
To address this, IE-PINN incorporates a dedicated strain network, which decouples and predicts strain, thereby significantly reducing the sensitivity in second derivatives. The discrepancy between strains derived from displacement and those directly predicted by strain network is minimized through a strain discrepancy loss term.
This approach substantially improves the stability and accuracy of elasticity estimation, even under noisy conditions. Figure \ref{fig: 2_1_2_dragon_dog_prediction} depicts the predictions made by the IE-PINN for all relevant quantities, including denoised displacements, strains, stresses, and elasticity parameters. These predictions were trained using noisy displacement data, and their corresponding error maps are shown in Figure \ref{Supple_errormap} in the Supporting Information.

\begin{figure}[t!]
    \centering
    \begin{minipage}{0.52\textwidth}
        \centering
        \includegraphics[width=\textwidth]{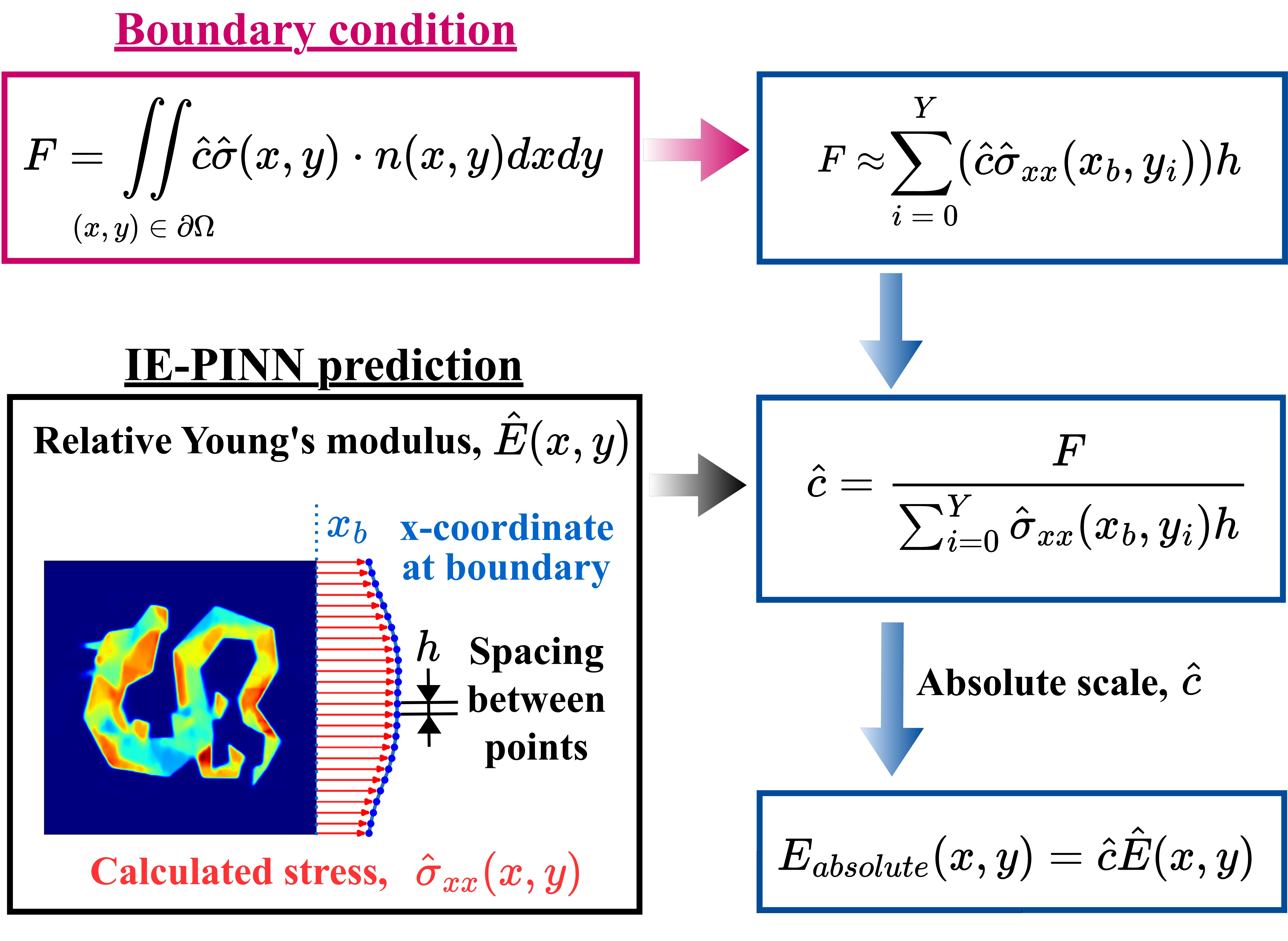}
        \caption{ Young's modulus scale calibration procedure. Upon completion of Phase 1 training, the IE-PINN provides a spatially varying relative Young's modulus field. In Phase 2, the absolute scale is calibrated by incorporating the known loading boundary conditions. Specifically, the predicted boundary stress from the relative Young's modulus is used to compute the resultant force, which is then aligned with the experimentally applied loading force to recover the true scale of Young's modulus.
        }
        \label{fig:calibration}
    \end{minipage}
    \hfill
    \begin{minipage}{0.44\textwidth}
        \centering
        \includegraphics[width=\textwidth]{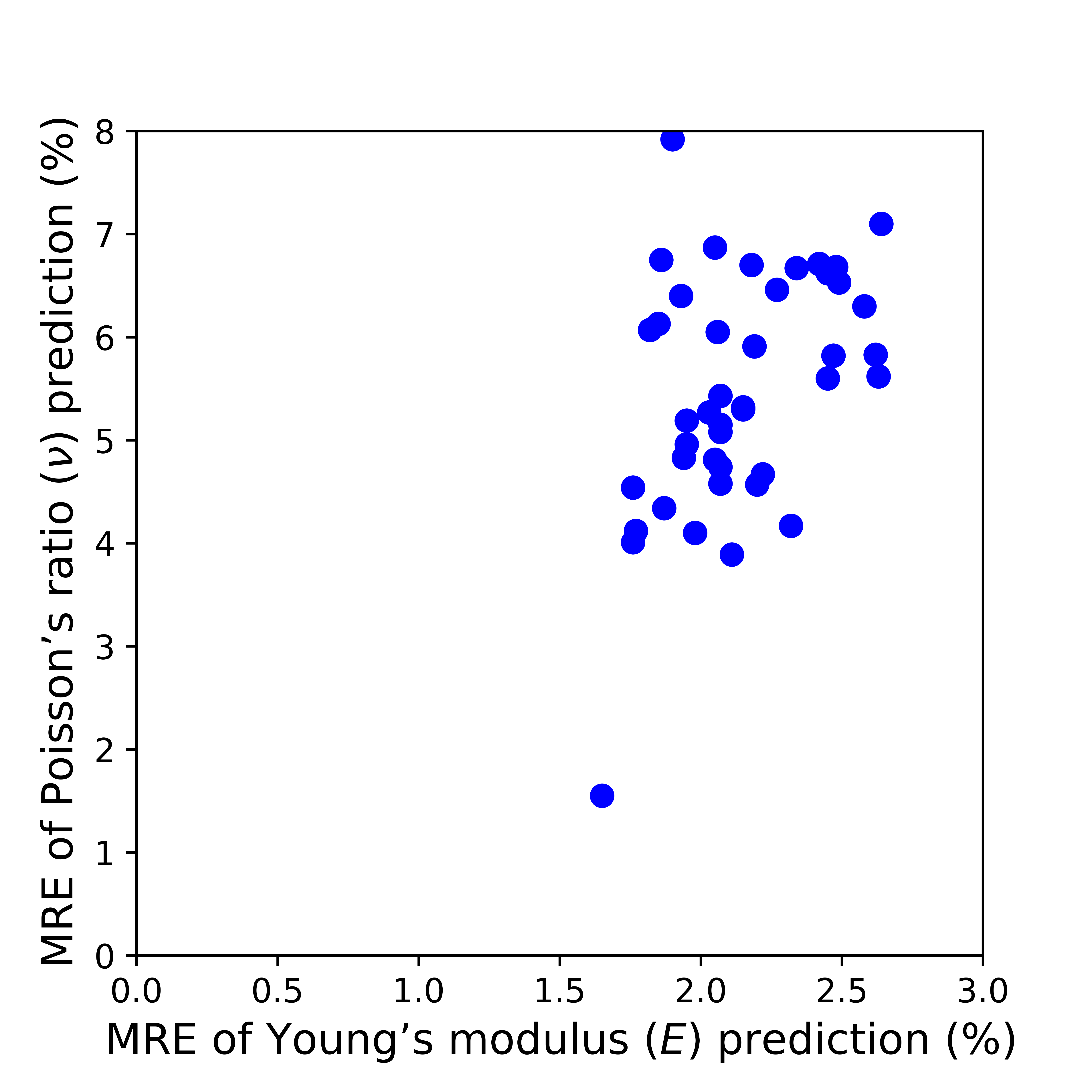}   
        \caption{The prediction error of Young’s modulus and Poisson’s ratio. The proposed model achieves a significantly low and consistently reliable mean relative error (MRE) across 50 independent datasets with noisy displacement data at a signal-to-noise ratio (SNR) of 1000, demonstrating robust accuracy in estimating the elasticity parameters.
        }
        \label{fig:2_1_4_prediction_error}
    \end{minipage}
\end{figure}

Another significant innovation of IE-PINN is its method for estimating absolute elasticity scales.
Accurate absolute elasticity estimation requires precise enforcement of boundary conditions related to the applied loading force. Directly incorporating these boundary conditions into PINN loss functions often causes ill-conditioned optimization and training failure.\cite{Cao2025}
To address this challenge, we calibrate the absolute scale of Young's modulus by aligning the boundary force, which is computed from the predicted stress based on the relative Young's modulus estimated from Phase 1, with the experimentally measured loading force, which is often available in practice.\cite{Rho1993, tai2007nanoscale, zhou2020elastic}
Figure \ref{fig:calibration} illustrates the calibration procedure employed in this study using numerical integration of the predicted relative stress under the applied loading force, where the technical details are described in Section \ref{ssec:s_formualtion_CalibrationE}.
The proposed two-step approach enables effective estimation of the elasticity distributions at the correct absolute scale, incorporates the boundary conditions during training, and maintains training stability. 

This research applies the proposed IE-PINN to a thin plate scenario under plane stress conditions, predicting the stress distributions ($\sigma_{xx}$, $\sigma_{yy}$, and $\tau_{xy}$) using constitutive and strain-displacement equations, rather than assuming uniform distributions. 
The model minimizes a combined total loss function comprising displacement data, strain discrepancy, PDE residual, and mean modulus constraint losses during training.
Each loss term is designed to reduce a specific source of errors: The displacement loss penalizes deviations between predicted and observed displacements; The strain discrepancy loss reduces the differences between the strain predictions from the strain network and those derived by differentiating the predicted displacements; The PDE residual loss enforces the governing equations by minimizing the PDE residuals computed from the strain and elasticity networks. 
Finite difference is employed for numerical differentiation. 
After training convergence, the IE-PINN reliably predicts displacement, strain, stress, Young’s modulus, and Poisson’s ratio distributions. 

IE-PINN maintains excellent predictive accuracy, whereas Elastnet demonstrates significant estimation failure under the same noisy condition (see Section \ref{ssec:displacement fitting}).
We further evaluate the performance of our proposed method across multiple datasets with different spatial elasticity distributions in Figure \ref{fig:2_1_4_prediction_error}. 
In the following sections, we systematically demonstrate the advantages of each component of IE-PINN, including the neural network architecture (Sections \ref{ssec:displacement fitting} and \ref{ssec:positional and activation}), robustness to varying noise levels (Section \ref{ssec:noise sensitivity}), robustness of the calibration procedure to the mean modulus constraint, and the impact of pretraining (Section \ref{ssec:pretraining}).

\subsection{Advantages of Displacement Fitting and Decoupled Strain Prediction} \label{ssec:displacement fitting}
\begin{figure}[h]
    \centering
    \begin{subfigure}[t]{0.49\textwidth}
        \centering
        \includegraphics[width=\textwidth]{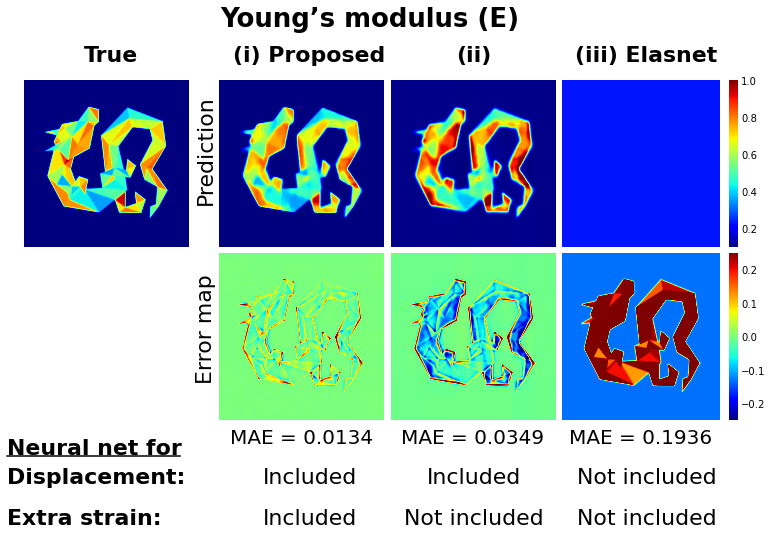} 
        \caption{Estimated Young's modulus across different models. 
        }
        \label{2_4_1_E}
    \end{subfigure}
    \hfill
    \begin{subfigure}[t]{0.49\textwidth}
        \centering
        \includegraphics[width=\textwidth]{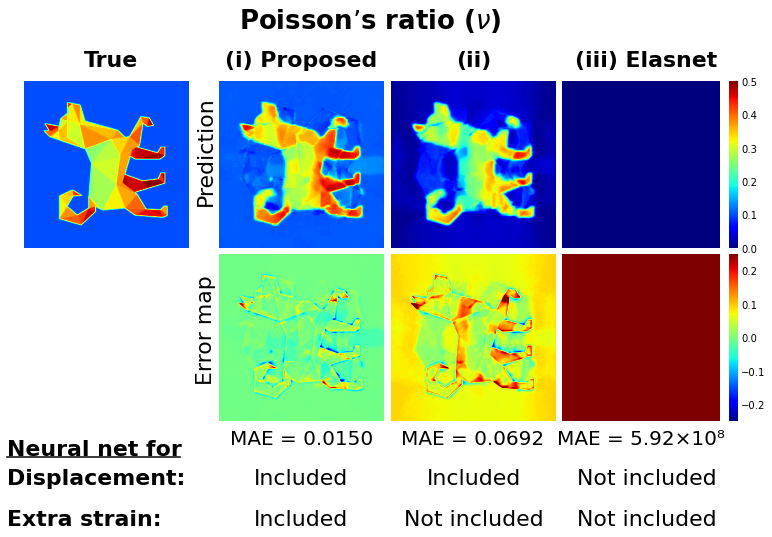} 
        \caption{Estimated Poisson’s ratio across different models. 
        }
        \label{2_4_1_v}
    \end{subfigure}
    \caption{Young's modulus ($E$) and Poisson’s ratio $(\nu)$ predictions and corresponding error maps were obtained from different models. All models were trained using the same noisy displacement data with a signal-to-noise (SNR) ratio of 1000. (i) IE-PINN (Proposed) incorporates both displacement and strain networks. (ii) Model with only a displacement network. (iii) Elastnet does not employ any functional approximation to fit displacement or strain.\cite{Chen2023} Elastnet fails to learn meaningful Young's modulus distribution when displacement data are noisy. Employing Function approximation over noisy displacement data tends to denoise the measurements and improve the stability. Additionally, incorporating a dedicated strain network enhances the robustness and accuracy of elasticity estimation.
    \label{fig:Performance:E_v_Models}
    }
\end{figure}

For robust elasticity estimation from noisy displacement data, using a neural network fitted to displacement measurements is essential. 
Figure \ref{fig:Performance:E_v_Models}(a) and \ref{fig:Performance:E_v_Models}(b) compare the estimation and error maps of Young's modulus and Poisson's ratio from our proposed IE-PINN model against two benchmarks. Elastnet  directly applies finite differentiation to the (noisy) displacement measurements and the stresses calculated therefrom.\cite{Chen2023} Direct differentiation significantly amplifies the errors in the noise, making the elasticity estimation vulnerable to noise.
Figure \ref{fig:Performance:E_v_Models}(a)(iii) and \ref{fig:Performance:E_v_Models}(b)(iii) demonstrate the detrimental impacts of noise to the Elastnet, failing in the estimation of both Young's modulus and Poisson's ratio.
Fitting displacement data alone can mitigate the adverse effects of noise to some extent. Figure \ref{fig:Performance:E_v_Models}(a)(ii) and \ref{fig:Performance:E_v_Models}(b)(ii) demonstrate that both Young's modulus and Poisson's ratio are estimated with some errors. Nonetheless, the displacement network brings another challenge in that it is sensitive to its second derivative. Data fitting errors propagated to the second derivative make the elasticity estimation unstable. As a result, in other datasets, solely using the displacement neural network also often fails in elasticity estimation (Figure \ref{Supple_RabbitRoaster_SNR1000}(a)(ii) and \ref{Supple_RabbitRoaster_SNR1000}(b)(ii) in Supporting Information).
Figure \ref{fig:Performance:E_v_Models}(a)(i) and \ref{fig:Performance:E_v_Models}(b)(i) present our proposed method, which reduces the estimation errors significantly in both Young's modulus and Poisson's ratio. The estimated values are quite close to the true ones.
The strain network not only improves the accuracy of elasticity estimation but also stabilizes the elasticity estimation significantly under the noisy conditions. 
Under high noise conditions, the displacement network alone is not sufficient. In contrast, the proposed IE-PINN successfully estimates both Young's modulus and Poisson's ratio, confirming its robustness against noise (Figure \ref{Supple_DragonDog_SNR500} and \ref{Supple_DragonDog_SNR100} in the Supporting Information).

\subsection{Robustness to Displacement Noise} \label{ssec:noise sensitivity}
\begin{figure}[t!]
    \centering
    \begin{subfigure}[t]{0.49\textwidth}
        \centering
        \includegraphics[width=\textwidth]{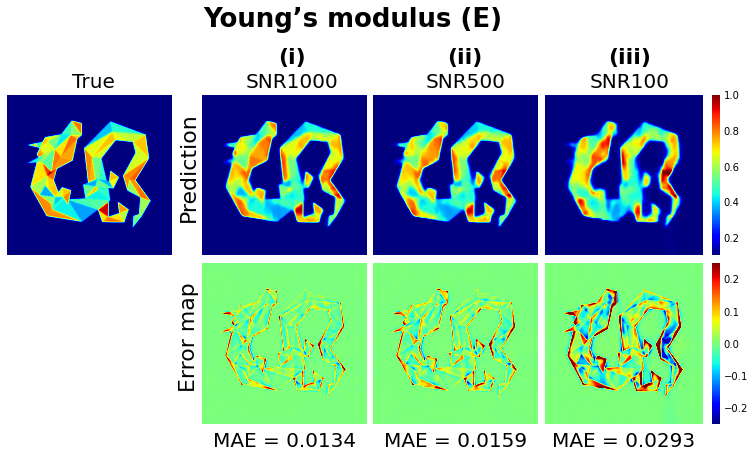} 
        \caption{Estimated Young's modulus across noise levels. 
        }
        \label{2_2_1_SNR_E}
    \end{subfigure}
    \hfill
    \begin{subfigure}[t]{0.49\textwidth}
        \centering
        \includegraphics[width=\textwidth]{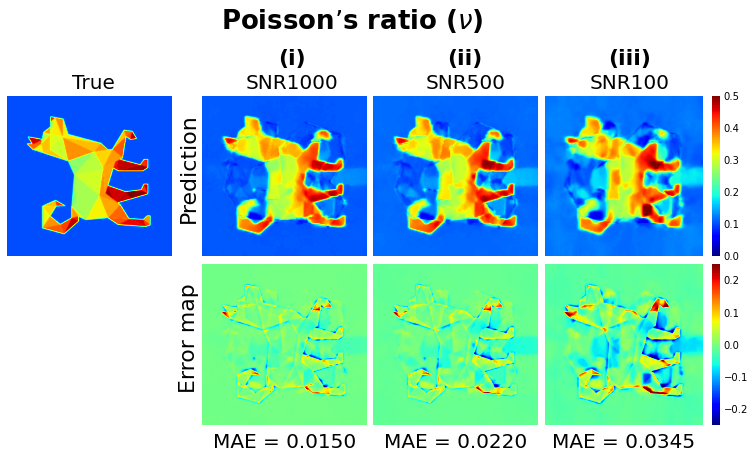} 
        \caption{Estimated Poisson’s ratio across noise levels. 
        }
        \label{2_2_1_SNR_v}
    \end{subfigure}
    \caption{Predicted (a) Young's modulus ($E$) and (b) Poisson’s ratio $(\nu)$, along with their corresponding error maps, evaluated across varying noise levels (signal-to-noise ratio, SNR). IE-PINN was trained using the same displacement data across three different SNRs: (i) SNR = 1000,  (ii) SNR = 500, and (iii) SNR = 100. Although the prediction errors increase as the noise level rises, the model performance remains robust across noise levels.
    \label{fig:Performance:E_v_SNR}
    }
\end{figure}

\begin{figure}[t!]
    \centering
    \begin{minipage}[t]{0.49\textwidth}
        \centering
        \includegraphics[width=\textwidth]{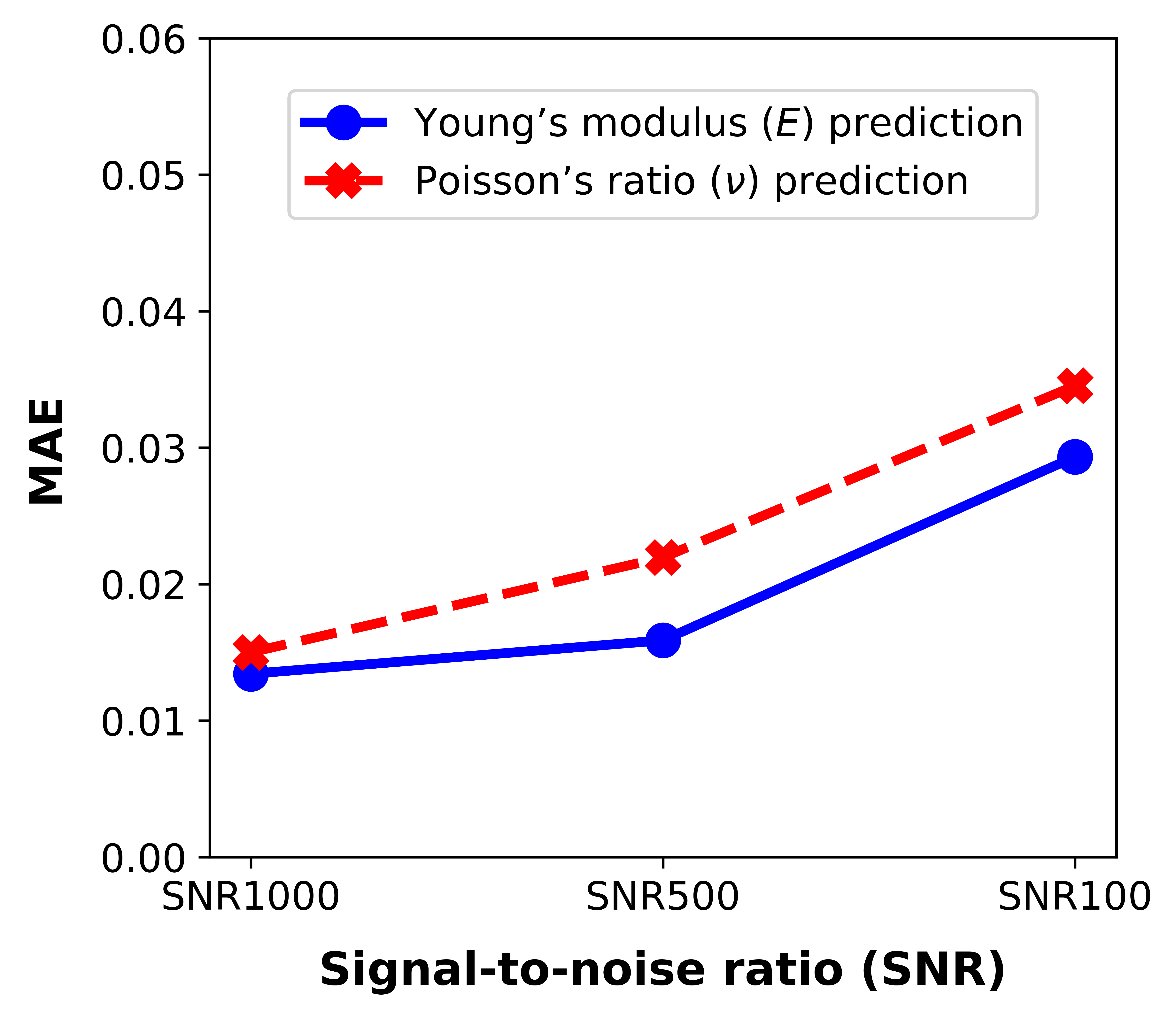} 
        \caption{Robustness of prediction errors across different noise levels. The predicted Young's modulus and Poisson's ratio exhibit strong robustness to noise, with only minimal degradation in accuracy even at higher noise levels.}
        \label{2_2_1_SNR}
    \end{minipage}
    \hfill
    \begin{minipage}[t]{0.49\textwidth}
        \centering
        \includegraphics[width=\textwidth]{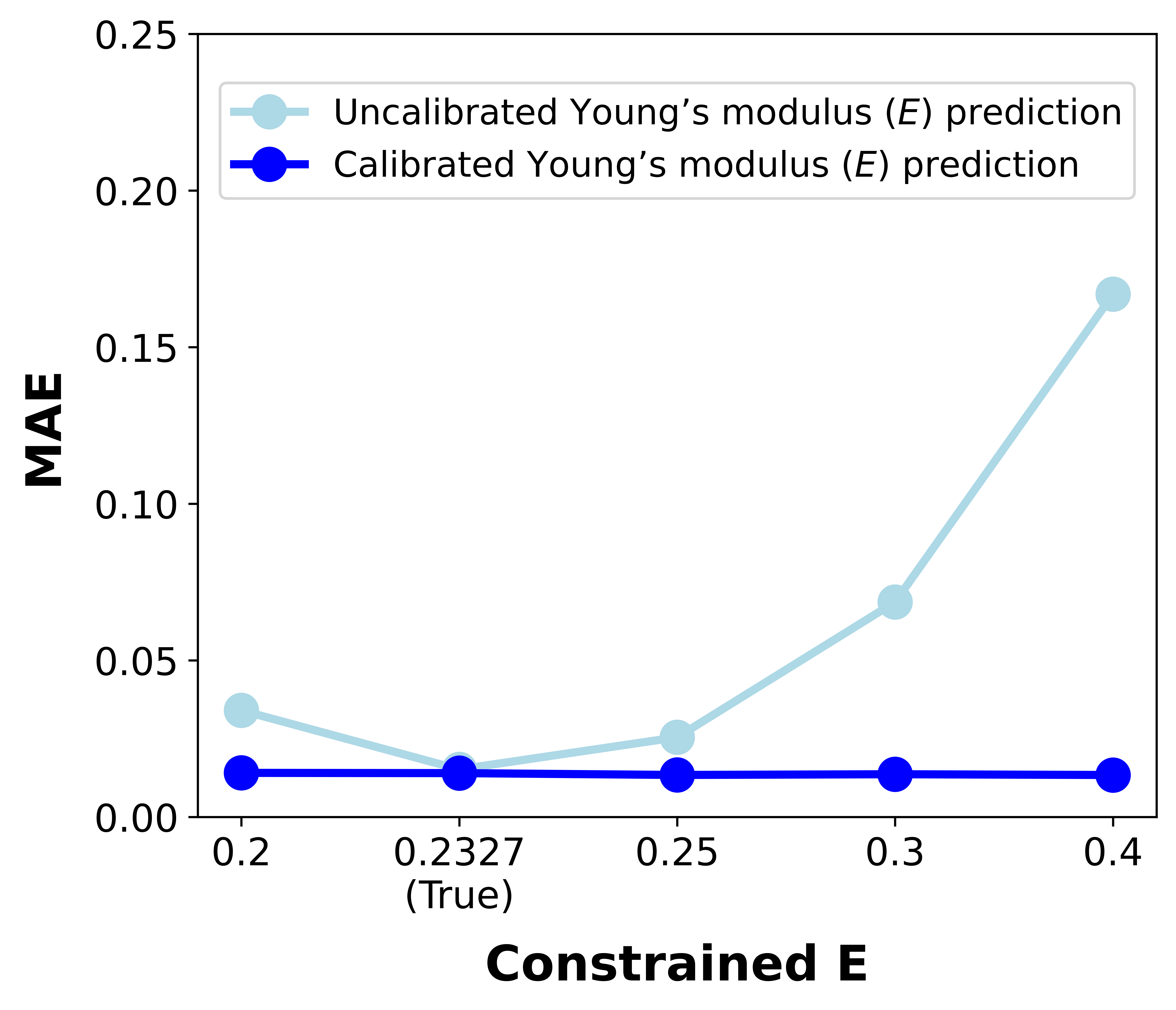} 
        \caption{Impact of constrained mean Young’s modulus. Although the model is trained with various values of constrained mean Young’s Modulus, the proposed absolute scale calibration technique effectively adjusts the predicted relative Young's modulus from IE-PINN in Phase 1 to their corresponding absolute scales. The prediction errors remain consistent across different constrained mean values, demonstrating the robustness of the calibration approach.}
        \label{fig:MAE over mean E}
    \end{minipage}
\end{figure}

Noise in displacement data poses a significant challenge for accurate elasticity estimation. 
As aforementioned, the noises in displacement are amplified through differentiation in the calculation of strain and equilibrium equation, worsening the accuracy of elasticity estimation. 
To study the sensitivity of the IE-PINN to noise in the displacement data, we investigate different noise levels based on three signal-to-noise ratios (SNR), the ratios of mean displacement over the standard deviation of errors: 1000, 500, and 100. The predicted field and error map for Young’s modulus and Poisson’s ratio are presented in Figure \ref{fig:Performance:E_v_SNR}(a) and  Figure \ref{fig:Performance:E_v_SNR}(b), respectively. The mean absolute errors (MAE) in the IE-PINN prediction of Young's modulus and Poisson’s ratio across noise levels are shown in Figure \ref{2_2_1_SNR}. The results demonstrate that the prediction errors remain low across different noise levels, highlighting the model's robustness in elasticity estimation. The MAE for Young’s modulus at SNR 1000 is not significantly different from that at SNR 500 (Figure \ref{Supple_DragonDog_SNR500} in Supporting Information). When the noise level is increased tenfold (SNR 100), the prediction errors for both Young’s modulus and Poisson’s ratio show a noticeable increase (Figure \ref{Supple_DragonDog_SNR100} in Supporting Information).

\subsection{Absolute Scale Calibration and Mean Young's Modulus Constraint} \label{s_constrained_meanE}
The boundary condition on loading force is critical as it determines the unit of Young's modulus. Without this boundary condition, only the relative distribution can be obtained. 
However, integrating the boundary conditions directly into the loss functions of PINN often results in ill-conditioned optimization problems, which can cause training difficulties or failures. 
Previous methodologies typically assume prior knowledge of the true mean Young’s modulus,\cite{Chen2021, Chen2023} or stress distributions,\cite{Kamali2023,Kamali2024} which is generally unknown in practical applications, complicating absolute scale elasticity estimation. 
To overcome this limitation, IE-PINN initially estimates Young's modulus with an arbitrary mean value, generating a relative modulus distribution. Subsequently, IE-PINN employs a novel calibration method to recover the absolute scale by aligning the predicted relative boundary stress distribution with experimentally measured loading forces as illustrated in Figure \ref{fig:calibration}.

Figure \ref{fig:MAE over mean E} demonstrates the impacts of various arbitrary mean constraints on the prediction errors before and after scale calibration (Prediction accuracy is included in Table \ref{Supple_table_meanE} in Supporting Information).
Initially, high MAE is observed before calibration due to inaccurate mean elasticity assumptions. Nevertheless, the proposed calibration technique successfully identifies the correct scale from boundary stress predictions, achieving performance comparable to the ideal case.
The figure confirms that the calibration performance is significantly robust against the constrained mean modulus values. 
Additionally, Figure \ref{Supple_ConstrainedMeanE} in Supporting Information shows that Poisson’s ratio prediction is consistently robust regardless of the imposed mean constraints. 
This independent calibration scheme maintains precise absolute-scale Young's modulus estimation while preserving the training stability of IE-PINN stability.

\subsection{Positional Encoding and Activation Functions}
\label{ssec:positional and activation}
Neural networks with low-dimensional positional inputs (coordinates $x$ and $y$) often struggle to represent complex functions,\cite{Sitzmann2006, Mildenhall2021} resulting in ill-conditioned optimization and uninformative gradients.
To address this, we integrate a positional encoding function technique,\cite{Vaswani2017} transforming the two-dimensional coordinates into a richer multidimensional latent representation. This significantly improves the accuracy of elasticity estimations (Figure \ref{Supple_ImportancePositionalEncodingfunction} and Table \ref{Supple_table_positional} in Supporting Information).

\begin{figure}[h]
    \centering
    \begin{minipage}[t]{0.45\textwidth}
        \centering
        \includegraphics[width=\linewidth]{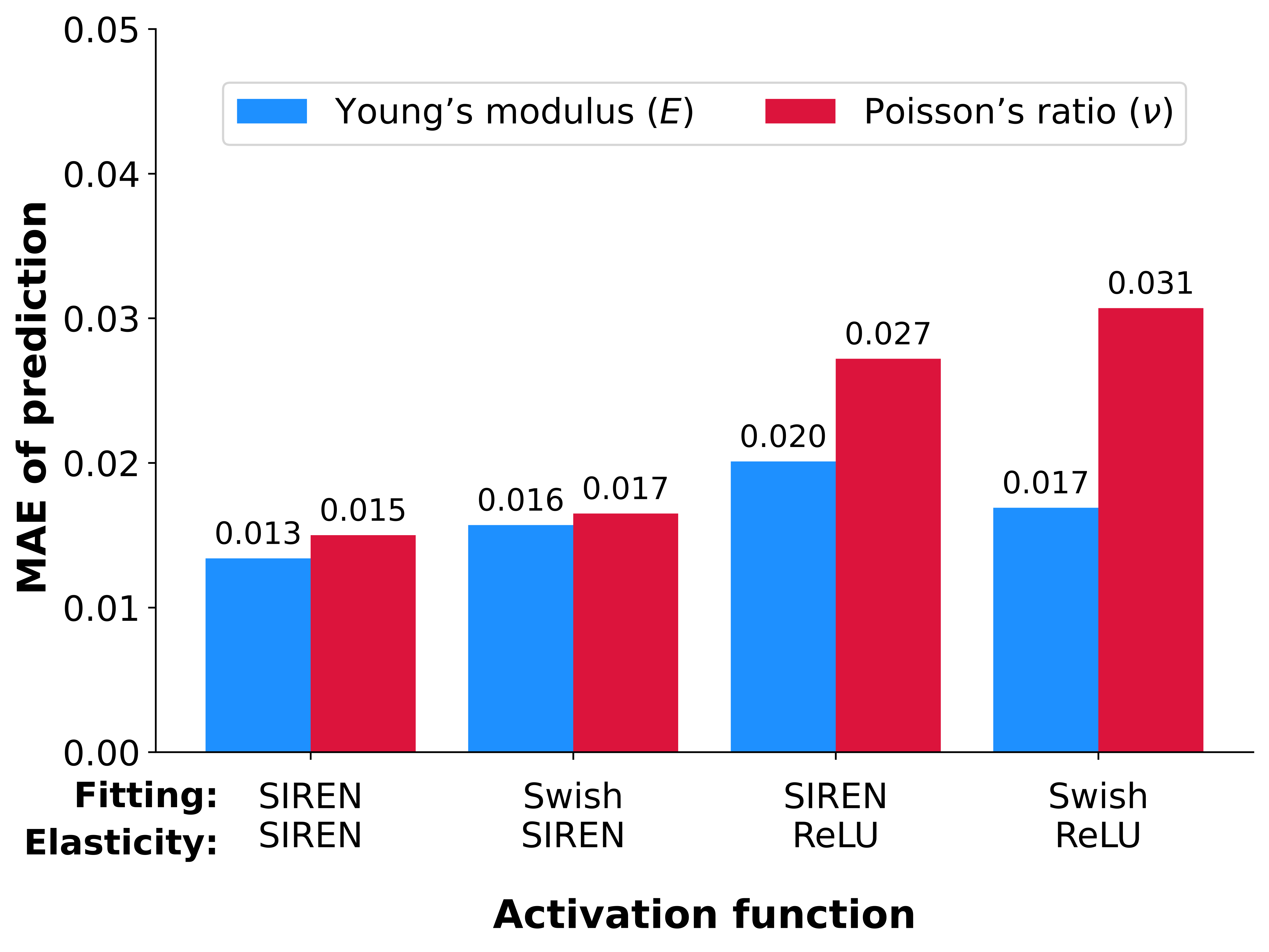} 
        \caption{Performance comparison across different activation functions. The prediction errors in Young's modulus (E) and Poisson’s ratio $(\nu)$ are presented for models using various activation functions in neural networks (\textbf{Fitting} denotes both displacement and strain networks, \textbf{Elasticity} denotes elasticity network). Among the activation functions evaluated, the SIREN activation function achieved the highest accuracy across all elastic property predictions.\cite{Sitzmann2020}}
        \label{2_5_1_fittingModel}
    \end{minipage}
    \hfill
    \begin{minipage}[t]{0.45\textwidth}
        \centering
        \includegraphics[width=\linewidth]{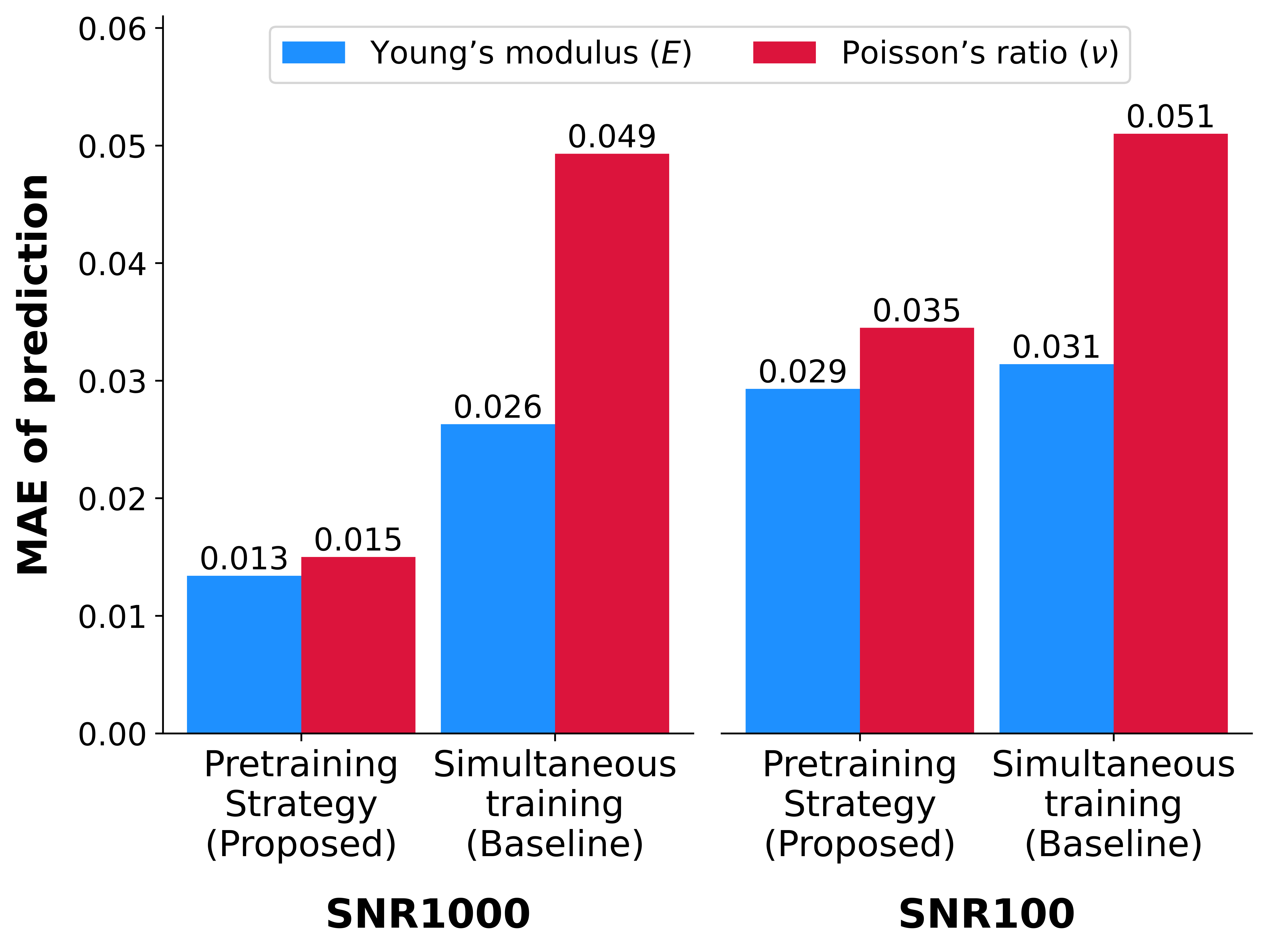} 
        \caption{Impact of pretraining strategy. The prediction errors in Young's modulus ($E$) and Poisson’s ratio $(\nu)$ are compared between two training strategies: A pretraining scheme that sequentially trains neural networks versus simultaneous training. Models are evaluated using noisy displacement data under two levels, corresponding to SNR of 1000 and 100. The results demonstrate that employing a pretraining strategy significantly improves the prediction accuracy of both elasticity parameters, achieving up to 50\% reduction in error.}
        \label{2_6_1_Pretraining}
    \end{minipage}
\end{figure}

Another critical aspect influencing neural network performance in PINN is the choice of activation functions. 
Traditional activation functions often suffer from vanishing/exploding gradients during training, reducing training efficiency.
IE-PINN employs the sine activation function (SIREN),which has proven effective in precise gradient and divergence prediction.\cite{Sitzmann2020}
Figure \ref{2_5_1_fittingModel} compares elasticity estimation errors across different activation functions, specifically evaluating the SIREN with Swish, used in Elastnet.\cite{Chen2023}
The same activation functions are used for displacement and strain fitting (denoted as Fitting) and those for Young's modulus and Poisson's ratio (denoted as Elasticity). 
Results demonstrate that using SIREN consistently produces the lowest MAEs for both Young's modulus and Poisson's ratio estimations.

\subsection{Pretraining Strategy}\label{ssec:pretraining}
Training PINN is significantly more challenging than training conventional neural networks due to the complexity introduced by multiple interconnected networks and loss terms. IE-PINN involves three distinct neural networks, each with specific loss terms, and their roles in solving the linear elasticity PDEs, including displacement fitting, strain discrepancy, and PDE residual losses.
To enhance the training efficacy, IE-PINN employs a sequential pretraining strategy. Initially, the displacement network is trained independently using the noisy displacement data (for 50,000 iterations) with only the data loss, which is a conventional neural network training. Subsequently, the strain network is trained additionally along with the displacement network by minimizing discrepancies with strains derived from the displacement network (for 100,000 iterations). 
Finally, all IE-PINN networks are trained using all loss terms. 
Figure \ref{2_6_1_Pretraining} compares the performance of IE-PINN with and without pretraining strategy under different noise levels (SNR of 1000 and 100) under the same total number of training iterations. The results demonstrate that pretraining effectively improves prediction accuracy, confirming its effectiveness in both low and high noise levels. 
Further details are provided in Figure \ref{Supple_Pretraining} of Supporting Information.

\subsection{Related Work and Future Study}\label{ssec:relatedwork}
Several studies have explored inverse physics-informed neural networks for elasticity estimation. Early approaches primarily considered homogeneous materials with constant elasticity assumptions; some relied on both displacement and stress measurements, \cite{Haghighat2021, Sahin2024, Mallampati2021, Min2024} while others used only displacement data \cite{ Gao2022,Kianian2024, Hamel2022}. 
More recent efforts have extended to heterogeneous materials for spatially varying elasticity estimation.
Some studies assume incompressible material behavior and estimate only Young's modulus while assuming Poisson's ratio constant \cite{Chen2021,Ragoza2023}.
Due to the ineffective performance after integrating the boundary condition on loading force, some works exclude the boundary conditions and obtain relative Young's modulus distribution rather than absolute scale.\cite{Ragoza2023, Chen2023} The prior knowledge of the true mean Young's modulus can be integrated as another loss to obtain the accurate distribution of elasticity distributions \cite{Chen2021, Chen2023}.
Other methods use strain data under the assumption of known boundary stress distribution, which is typically unavailable in real-world scenarios \cite{Kamali2023, Kamali2024}.

Elastnet uses the strain data for incompressible materials in the earlier model,\cite{Chen2021} and displacement data for compressible materials in the more recent version,\cite{Chen2023} employing finite-difference approximations to estimate elasticity. Unlike typical PINN-based methods, Elastnet does not fit a functional model (e.g., neural networks); instead, it directly applies numerical differentiation to all the variables, including noisy measurements, making it particularly sensitive to noise, as discussed in Section \ref{ssec:displacement fitting}. 
Additionally, boundary conditions are not incorporated into their estimation, resulting in estimated Young's modulus being expressed in relative scales rather than absolute ones. 
In contrast, our proposed IE-PINN demonstrates robustness to measurement noises and enables estimation of heterogeneous Young's modulus on an absolute scale, based on the externally applied force that can be experimentally measured using localized mechanical testing techniques such as ultrasound,\cite{Rho1993} nanoindentation,\cite{tai2007nanoscale} and atomic force microscopy.\cite{zhou2020elastic}

Several gaps in the current research remain to be addressed in future work. In many clinical applications, the measurement data are of low spatial resolution, making precise elasticity estimation particularly challenging. Additionally, the extension to three-dimensional (3D) elasticity estimation remains an open research direction. The 3D displacement data experimentally measured using the DVC method often contains measurement noise, further complicating the problem. Advancing 3D inverse elasticity estimation would benefit a broad range of applications, including disease diagnosis through biomedical imaging, materials design in manufacturing, and structural analysis in construction, such as detecting internal defects or optimizing the behavior of composite materials under complex loading conditions.

\section{Conclusion}
In this work, we present an Inverse Elasticity Physics-Informed Neural Network (IE-PINN) model that effectively estimates heterogeneous Young's modulus and Poisson's ratio from noisy displacement data by addressing the challenges posed by measurement noise and the difficulty of recovering Young's modulus at the correct absolute scale. 
IE-PINN employs three distinct neural networks to stabilize the estimation, each predicting displacements, strains, and elasticity parameters. 
Specifically, the extra strain network significantly stabilizes elasticity estimation under noisy data. 
This novel architecture achieves strong robustness against noise in displacement and accurately reconstructs spatially varying elastic properties. 
IE-PINN also incorporates additional strategies: a positional encoding function, a sine activation function, and a pretraining strategy. 
We propose a two-phase approach: Phase 1 estimates the spatial distribution of Young's modulus in relative scales alongside Poisson's ratio. Phase 2 recovers the correct absolute scale of Young's modulus using loading boundary conditions.
IE-PINN robustly achieves absolute-scale elasticity estimation from noisy displacements, whereas state-of-the-art methods exhibit significant performance degradation under noisy data. 
Our proposed approach is particularly advantageous for clinical elastography, material design, and structural optimization, where displacements are measured in response to defined external loading conditions and often contain noise arising from equipment or environmental factors.

\section{Experimental Section} \label{s_experimental_section}
\subsection{Integration of Domain Knowledge of Linear Elasticity}
Linear elasticity is a mathematical model in PDE that describes how solid objects deform and become internally stressed by prescribed loading conditions. 
IE-PINN seamlessly incorporates the physics governing mechanical behavior in linear elasticity.
We consider a two-dimensional setting with isotropic material and plane stress.
In IE-PINN, the numerical differentiation is performed by finite differentiation using a specially designed convolution kernel.\cite{Chen2021}
The estimation of elasticity parameters is performed by considering the entire PDE of linear elasticity. 

The deformation of an object is measured as the displacement field $\bm{u}=[u_x, u_y]^T$, a vector field that describes the movement of each material point from its original position, where ${u}_{x}$ and ${u}_{y}$ are the horizontal and vertical displacements. 
The strain vector 
is obtained using the gradient of the displacement field and captures both normal and shear deformation components as,
\begin{equation}
    \bm{\varepsilon} = 
    \begin{bmatrix}
        \varepsilon_{xx} \\
        \varepsilon_{yy} \\
        \gamma_{xy} \\
    \end{bmatrix}
    =
    \begin{bmatrix}
        \frac{\partial }{\partial x} & 0\\
        0 & \frac{\partial }{\partial y} \\
        \frac{\partial }{\partial y}  & \frac{\partial }{\partial x} \\
    \end{bmatrix}
    \begin{bmatrix}
        u_{x} \\
        u_{y} \\
    \end{bmatrix}\label{eq:displacementStrainRelation}
\end{equation}
where $\varepsilon_{xx}$ and $\varepsilon_{yy}$ are axial strains in the x- and y- directions, $\gamma_{xy}$ is shear strain, and ${\partial }/{\partial x}$ and ${\partial }/{\partial y}$ denote partial derivatives to the function with respect to $x$ and $y$, respectively.
In the IE-PINN framework, the strain vector $(\bm{\hat{\varepsilon}} = [\hat{\varepsilon}_{xx},\hat{\varepsilon}_{yy},\hat{\gamma}_{xy}]^T)$ is obtained by the derivative of the prediction of the displacement network at each positional input $(\bm{\hat{u}} = [\hat{u}_x,\hat{u}_y]^T)$ through finite differentiation. The strain at position $(i,j)$ is calculated as follows.
\begin{equation}
   \hat{\varepsilon}_{xx}(i,j) = \sum_{a=1}^2 \sum_{b=1}^2  {w_{x}}(a,b)\hat{u}_x(i+a-1,j+b-1) 
   \label{eq:strain_xx}
\end{equation}
\begin{equation}
   \hat{\varepsilon}_{yy}(i,j) = \sum_{a=1}^2 \sum_{b=1}^2  {w_{y}}(a,b)\hat{u}_y(i+a-1,j+b-1) 
   \label{eq:strain_yy}
\end{equation}
\begin{equation}
   \hat{\gamma}_{xy}(i,j) = \sum_{a=1}^2 \sum_{b=1}^2  {w_{y}}(a,b)\hat{u}_x(i+a-1,j+b-1) + {w_{x}}(a,b)\hat{u}_y(i+a-1,j+b-1)  
   \label{eq:strain_xy}
\end{equation}
where the ${w_{x}}(i,j)$ and ${w_{y}}(i,j)$ are the convolution kernels that facilitate the finite differentiation of the function, defined as follows.
\begin{equation}
    {w_{x}} = \begin{bmatrix}
                            -0.5 & 0.5 \\
                            -0.5 & 0.5 \\
                            \end{bmatrix} ,
                    {w_{y}} = \begin{bmatrix}
                            0.5 & 0.5\\
                            -0.5 & -0.5\\
                            \end{bmatrix}  \notag
\end{equation}

The stress, denoted as $\bm{\sigma}$, is the quantity of force per specific point after the force is applied along the surface area. 
In linear elasticity, the relationship between stress and strain is characterized by Young's modulus $E$ and Poisson’s ratio $\nu$ as follows,
\begin{equation}
    \bm{\sigma} = 
    \begin{bmatrix}
        \sigma_{xx} \\
        \sigma_{yy} \\
        \tau_{xy} \\
    \end{bmatrix}
    =
    \frac{E}{1-\nu^2}
    \begin{bmatrix}
        1 & \nu & 0\\
        \nu & 1 & 0\\
        0 & 0 & \frac{1-\nu }{2} \\
    \end{bmatrix}
    \begin{bmatrix}
        \varepsilon_{xx} \\
        \varepsilon_{yy} \\
        \gamma_{xy} \\
    \end{bmatrix}
    \label{eq:elasticCon}
\end{equation}

The residual force (also called PDE residual) is defined with respect to stress as follows.
\begin{align}
    r_x &=\frac{\partial \sigma_{xx}}{\partial x} + \frac{\partial \tau_{xy}}{\partial y}
    \\
    r_y &=\frac{\partial \tau_{xy}}{\partial x} + \frac{\partial \sigma_{yy}}{\partial y}
\end{align}

The stress vector is subject to meet the static equilibrium equations, where the residual forces 
 are zero at every coordinate point:
\begin{align}
    r_x=0 \text{\quad and \quad} r_y=0
\end{align}
The stress vector is predicted by using the strain vector and elastic parameters predicted by strain and elasticity networks at given coordinates, where the PDE residuals are derived by finite differentiation using convolution operations and to be minimized toward zero.

At $(i,j)$th position, the residual forces are calculated by using predicted stress vectors as follows.
\begin{align}
    r(i,j) = & \sum_{a=1}^3 \sum_{b=1}^3 \{ {w_{xx}}(a,b)\sigma_{xx}(i+a-1,j+b-1) \notag \\
    & + {w_{yy}}(a,b)\sigma_{yy}(i+a-1,j+b-1) \notag \\
    & + {w_{xy}}(a,b)\tau_{xy}(i+a-1,j+b-1) \}/ht \label{eq:residualPDE}
\end{align}
where ${w_{xx}}$, ${w_{yy}}$, and ${w_{xy}}$ are the kernels defined for the finite differentiation, and $h$ and $t$ are the vertical and horizontal distances between neighboring displacement data points.

$r_x$ is obtained by using the kernels defined below.
\begin{equation}
    w_{xx} = \begin{bmatrix}
                            -1 & 0 & 1\\
                            -1 & 0 & 1\\
                            -1 & 0 & 1 \\
                            \end{bmatrix} ,
                    w_{yy} = \begin{bmatrix}
                            0 & 0 & 0\\
                            0 & 0 & 0\\
                            0 & 0 & 0 \\
                            \end{bmatrix} ,
                    w_{xy} = \begin{bmatrix}
                            1 & 1 & 1\\
                            0 & 0 & 0\\
                            -1 & -1 & -1 \\
                            \end{bmatrix} \label{eq:kernelResidual_x}
\end{equation}
and $r_y$ is obtained by the kernel below.
\begin{equation}
    w_{xx} = \begin{bmatrix}
                            0 & 0 & 0\\
                            0 & 0 & 0\\
                            0 & 0 & 0 \\
                            \end{bmatrix} ,
                    w_{yy} = \begin{bmatrix}
                            1 & 1 & 1\\
                            0 & 0 & 0\\
                            -1 & -1 & -1 \\
                            \end{bmatrix} ,
                    w_{xy} = \begin{bmatrix}
                            -1 & 0 & 1\\
                            -1 & 0 & 1\\
                            -1 & 0 & 1 \\
                            \end{bmatrix} 
                            \label{eq:kernelResidual_y}
\end{equation}

\subsection{Inverse Elasticity Physic-Informed Neural Network}

Inverse elasticity physics-informed neural network estimates the spatial distributions of Young's modulus and Poisson's ratio that are expressed by neural networks. It is based on a hybrid approach integrating a data-driven neural network model with the physics law of elasticity. 
As described in Section \ref{sec:Result}, IE-PINN consists of three sets of neural networks, displacement, extra strain, and elasticity networks, describing the corresponding variables.
In training, IE-PINN aims to achieve four subgoals: (1) Fitting the displacement network to the displacement data, (2) Aligning Strain network prediction with the derived strains from the displacement neural network, (3) Constraining the average of Young's modulus to the given mean value, and (4) Satisfying the linear elasticity equilibrium equations \eqref{eq:residualPDE} while achieving the previous three objectives (1)-(3); Each subgoal can be achieved by minimizing the corresponding loss terms: displacement fitting loss $(\mathcal{L}_u)$, strain discrepancy loss $(\mathcal{L}_\varepsilon)$, mean modulus constraint loss $(\mathcal{L}_E)$, and the PDE residual loss $(\mathcal{L}_r)$.
The ultimate goal is to make the consisting neural networks achieve all four subgoals by minimizing these loss terms. One common way to solve such multi-objective optimization is to minimize the weighted sum of these losses in terms of neural network parameters: parameters of displacement network $\theta_u$, strain network $\theta_\varepsilon$, and elasticity network $\theta_E$.
The problem is then formulated as follows.
\begin{equation}
     \min_{\theta_u , \theta_\varepsilon , \theta_E} \Big[\lambda_{u} \mathcal{L}_{u}  + \lambda_{\varepsilon} \mathcal{L}_{\varepsilon} + \lambda_{r} \mathcal{L}_{r} +\lambda_{E} \mathcal{L}_{E}  \Big] 
\end{equation}
where $\lambda_{u}$ is the weight of displacement fitting loss function set at 2, $\lambda_{\varepsilon}$ is the weight of strain fitting loss function set at 1, $\lambda_{r}$ is the weight of residual of equilibrium loss function set at 3, and $\lambda_{E}$ is the weight of mean modulus constraint loss function set at 0.02. 

Each loss term is defined as the L1-norm of the average errors that deviate from the target values.
The displacement fitting loss $\mathcal{L}_{u}$ quantifies the average prediction errors of predicted displacement from the measurements.
\begin{equation}
    \mathcal{L}_{u}  = \frac{1}{N_xN_y} \sum_{i=1}^{N_x} \sum_{j=1}^{N_y} \left( | \hat{u}_x(i,j)- u^*_x(i,j) |+ | \hat{u}_y(i,j) - u^*_y(i,j) | \right) 
\end{equation}
where $N_x$ and $N_y$ are the total numbers of discrete points or pixels of the observed displacement field in the x- and y-directions, $\hat{u}_x$ and $\hat{u}_y$ are the predicted displacements in the x- and y-directions, ${u}^*_x$ and ${u}^*_y$ are the measured displacement in the x- and y-directions. 

The strain discrepancy loss measures the discrepancy between the predicted strain (from the strain network) and the strain derived from the predicted displacement of the displacement network through equation \eqref{eq:displacementStrainRelation} as follows.
\begin{equation}
    \mathcal{L}_{\varepsilon}  = \frac{1}{(N_x-1)(N_y-1)} \sum_{i=1}^{N_x-1} \sum_{j=1}^{N_y-1}  (| \hat{\varepsilon}_{xx}(i,j)- \varepsilon_{xx}(i,j) |+ | \hat{\varepsilon}_{yy}(i,j) - \varepsilon_{yy}(i,j) | 
      + | \hat{\gamma}_{xy}(i,j) - \gamma_{xy}(i,j) | ) 
\end{equation}
where $(N_x-1)$ and $(N_y-1)$ represent the reduced numbers of discrete points due to the convolution process as equation \eqref{eq:strain_xx},\eqref{eq:strain_yy} and \eqref{eq:strain_xy} with a reduction of 1, $\hat{\varepsilon}_{xx}$ and ${\hat{\varepsilon}}_{yy}$ are the predicted axial strains in the x- and y- directions, ${\varepsilon}_{xx}$ and ${\varepsilon}_{yy}$ are the axial strains derived from predicted displacement in the x- and y- directions, $\hat{\gamma}_{xy}$ is the predicted shear strain, and  ${\gamma}_{xy}$ is the shear strain derived from predicted displacement.

The physic laws are reflected through the residual loss of equilibrium equation as equation \eqref{eq:residualPDE}. The residual equilibrium loss is given by
\begin{equation}
    \mathcal{L}_{r}  = \frac{1}{(N_x-3)(N_y-3)} \sum_{i=1}^{N_x-3} \sum_{j=1}^{N_y-3} \frac{|r_x(i,j)|+|r_y(i,j)|}{\tilde{E}(i,j)}
\end{equation}
where  $(N_x-3)$ and $(N_y-3)$ represent the reduced numbers of discrete points, resulting from the convolution process to the stress in $(N_x-1) \times(N_y-1)$ using kernels in equation \eqref{eq:kernelResidual_x} and \eqref{eq:kernelResidual_y}, leading to dimension reduction of 2, $r_x(i,j)$ and $r_y(i,j)$ are the PDE residuals in \eqref{eq:residualPDE}, and $\tilde{E}(i,j)$ is the sum of the predicted Young's modulus values in a three by three sub-region, defined as.
\begin{equation}
    \tilde{E}(i,j)  =  \sum_{a=1}^{3} \sum_{b=1}^{3} {\hat{E}(i+a-1,j+b-1)}
\end{equation}

Determining the absolute scale of Young's modulus requires the boundary condition. For efficacy, IE-PINN estimates the relative modulus distribution in Phase 1 using an arbitrary mean modulus. The corresponding mean modulus constraint loss is added as follows:
\begin{equation}
    \mathcal{L}_{E}  =  \frac{1}{(N_x-1)(N_y-1)} \sum_{i=1}^{N_x-1} \sum_{j=1}^{N_y-1} \left|\hat{E}(i,j)-E_c \right|
\end{equation}
where $(N_x-1)$ and $(N_y-1)$ represent the reduced numbers of discrete points with the same dimension as predicted strain, $\hat{E}(i,j)$ is the predicted Young's modulus from elasticity network, and $E_c$ is the constrained mean value of Young's modulus. The role of this loss is to constrain the mean of prediction to a specific arbitrary $E$ value. Our calibration method can facilitate the applied force with the predicted stress from constrained mean Young's modulus value to recover the absolute Young's modulus.  
The training was done by Adam Optimizer with pretraining described in Section \ref{2_6_1_Pretraining}.

The accuracy of prediction among the same dataset is measured by the Mean Absolute Error (MAE), whereas the Mean Relative Error (MRE) is used to compare the prediction accuracy across different data sets. Both measures are defined as
\begin{equation}
    MAE = \frac{1}{(N_x-1)(N_y-1)} \sum_{i=1}^{N_x-1} \sum_{j=1}^{N_y-1} \left( | \hat{E}(i,j)- E(i,j) | \right) 
\end{equation}
\begin{equation}
    MRE = \frac{100}{(N_x-1)(N_y-1)} \sum_{i=1}^{N_x-1} \sum_{j=1}^{N_y-1} \left( \frac{| \hat{E}(i,j)- E(i,j) | }{E(i,j)}\right) 
\end{equation}
where $\hat{E}$ is a prediction of the quantity of interest, such as Young's modulus, displacement, strain, and Poisson's ratio. Moreover, $E$ is the measured or actual value of quantity.

\subsection{Young's Modulus Scale Calibration Technique} \label{ssec:s_formualtion_CalibrationE}

The inverse elasticity problem is generally ill-posed; estimation from displacement is unstable. The boundary condition is theoretically required to obtain the absolute value of Young's modulus. 
The existing methods use 1) known internal stress distributions, 2) known boundary stress distributions, or 
3) impose the true mean Young's modulus loss to guarantee a unique solution to the problem; however, the obtained solution is a relative Young's modulus with respect to the constrained mean of Young's modulus value. Moreover, the mean of Young's modulus is always unknown.   

The scale calibration technique is proposed to recover the absolute Young's modulus field from the relative stress field. This approach relies on determining the absolute scale of Young's modulus by utilizing the Neumann boundary condition known as the traction boundary condition \cite{Barber2009-kb, Sadd2014-fk}. This condition indicates the resultant of surface force $(\bm{F}_{\boundarySpace})$ over the entire boundary surface $\boundarySpace$ that can be expressed as the integral of a surface force density function $\bm{T}^n(x)$.
\begin{equation}
    \bm{F}_{\boundarySpace} = \int\int_{(x,y)\in {\boundarySpace}} \bm{T}^n(x,y)dxdy
\end{equation}
Commonly, the surface force density is referred to as the traction vector that varies on the spatial location. In this study that focuses on a thin rectangular plate with a plane stress assumption,
the geometry of the object is assumed to be a two-dimensional object where the traction is defined as below\cite{Sadd2014-fk}.
\begin{equation}
     \bm{T}^n(x,y) =   \begin{bmatrix}
                T^{(b)}_x(x,y) \\ T^{(b)}_y(x,y)
                \end{bmatrix} 
            =  \begin{bmatrix}
                \sigma^{(b)}_{xx}(x,y)n_x + \tau^{(b)}_{xy}(x,y)n_y\\ \tau^{(b)}_{xy}(x,y)n_x + \sigma^{(b)}_{yy}(x,y)n_y
                \end{bmatrix} 
\end{equation}
where $T^{(b)}_x$ and $T^{(b)}_y$ are the traction force at the boundary in the x- and y-directions, $\sigma^{(b)}_{xx}$ and $\sigma^{(b)}_{yy}$ are the stress at boundary in the x- and y-directions, $\tau^{(b)}_{xy}$ is a shear stress at boundary, $n_x$ and $n_y$ are normal vector to x- and y-directions

In the simulated dataset used in this study, a boundary force is applied in the x-direction, acting perpendicular to the right surface of the object.
A detailed description of the loading boundary condition is included in Supplementary Note S2 in Supporting Information.
The corresponding traction force per spatial coordinates is then as follows.
\begin{equation}
        T^n(X) =  T_x^{(b)}(x,y)  =
                \sigma_{xx}^{(b)}(x,y)n_x + \tau_{xy}^{(b)}(x,y)n_y
\end{equation}
As the applied force is perpendicular to the surface, it can be written as the vector form with $Fn_x+0n_y$. Then, the boundary force $F$ is equivalent to the total traction force at the boundary in the x-direction that can be calculated from the predicted stress at the boundary as
\begin{align}
    F =\iint_{(x,y)\in\boundarySpace} \sigma_{xx}^{(b)}(x,y)dx dy
\end{align}

However, the IE-PINN only predicts relative stress ${\sigma}_{xx}$ by using relative Young's modulus. Then, the boundary condition can be written with a multiplier $\hat{c}$ to recover the correct absolute scale of Young's modulus. 

\begin{align}
    F =\iint_{(x,y)\in\boundarySpace} \hat{c}\hat{\sigma}_{xx}^{(b)}(x,y) dx dy
\end{align}

By using numerical integration, the boundary force is obtained as follows.
\begin{align}
    F & \approx \sum_{i=0}^{Y}\hat{c}\hat{\sigma}_{xx}(x_b,y_i))h
\end{align}
The predicted relative Young's modulus from the model is eventually calibrated with the absolute scale to be the absolute Young modulus value. The formulation of the absolute scale for calibration is as
\begin{align}
   \hat{c}=\frac{F}{\sum^Y_{i=0}\hat{\sigma}_{xx}(x_b,y_i)h}
\end{align}
And the absolute Young modulus $E_{absolute}(x,y)$ can be tuned with the absolute scale ($\hat{c}$) to the predicted relative field from neural network ($\hat{E}(x,y)$) as:
\begin{align}
   E_{absolute}(x,y) & = \hat{c}\hat{E}(x,y)
\end{align}

\subsection{Neural Network Architecture}
All three neural networks (a displacement, a strain, and an elasticity network) in IE-PINN have 16 fully connected hidden layers with 128 neurons. The activation function between each fully connected hidden layer is a sine activation function named "SIREN".\cite{Sitzmann2006} Only the elasticity network has the last layer as the softplus activation function to guarantee the positivity of prediction. 
In all three networks, the input coordinates will pass through a positional encoding layer to achieve a more meaningful representation for learning. In this study, the sine and cosine functions with different frequencies,\cite{Vaswani2017} are used as the positional encoding function at the input layer defined as
\begin{align}
    \gamma(x,2i) &= \sin( f^{2i/\omega} x)  \quad \quad \quad ,i \in \{1,2,\dots , \omega\} 
    \label{eq:PE1}
    \\
    \gamma(x,2j+1) &= \cos( f^{2j/\omega} x)   \quad \quad \quad ,j \in \{1,2,\dots , \omega\} 
    \label{eq:PE2}
\end{align}
where $x$ is the $x$ coordinate, and the positional encoding in \eqref{eq:PE1} and \eqref{eq:PE2} is  also employed to $y$ positional input. $f$ is the minimum frequency set at 0.0001, and $\omega$ is the maximum number of frequencies set at 64 in this work.

\newpage

\newpage

\setcounter{page}{1}

\setcounter{figure}{0}

\renewcommand{\thefigure}{S\arabic{figure}}

\captionsetup[figure]{
  labelformat=simple,
  labelsep=colon,
  name={Supplementary Figure},
  labelfont=bf,
  textfont=normalfont,
  width=0.95\linewidth
}
\setcounter{table}{0}
\renewcommand{\thetable}{S\arabic{table}}
\captionsetup[table]{
  labelformat=simple,
  labelsep=colon,
  name={Supplementary Table},
  labelfont=bf,
  textfont=normalfont,
  width=0.95\linewidth
}
\section*{Supplementary Figures}
\begin{figure}[h]
    \centering
    \includegraphics[width=1.0\textwidth]{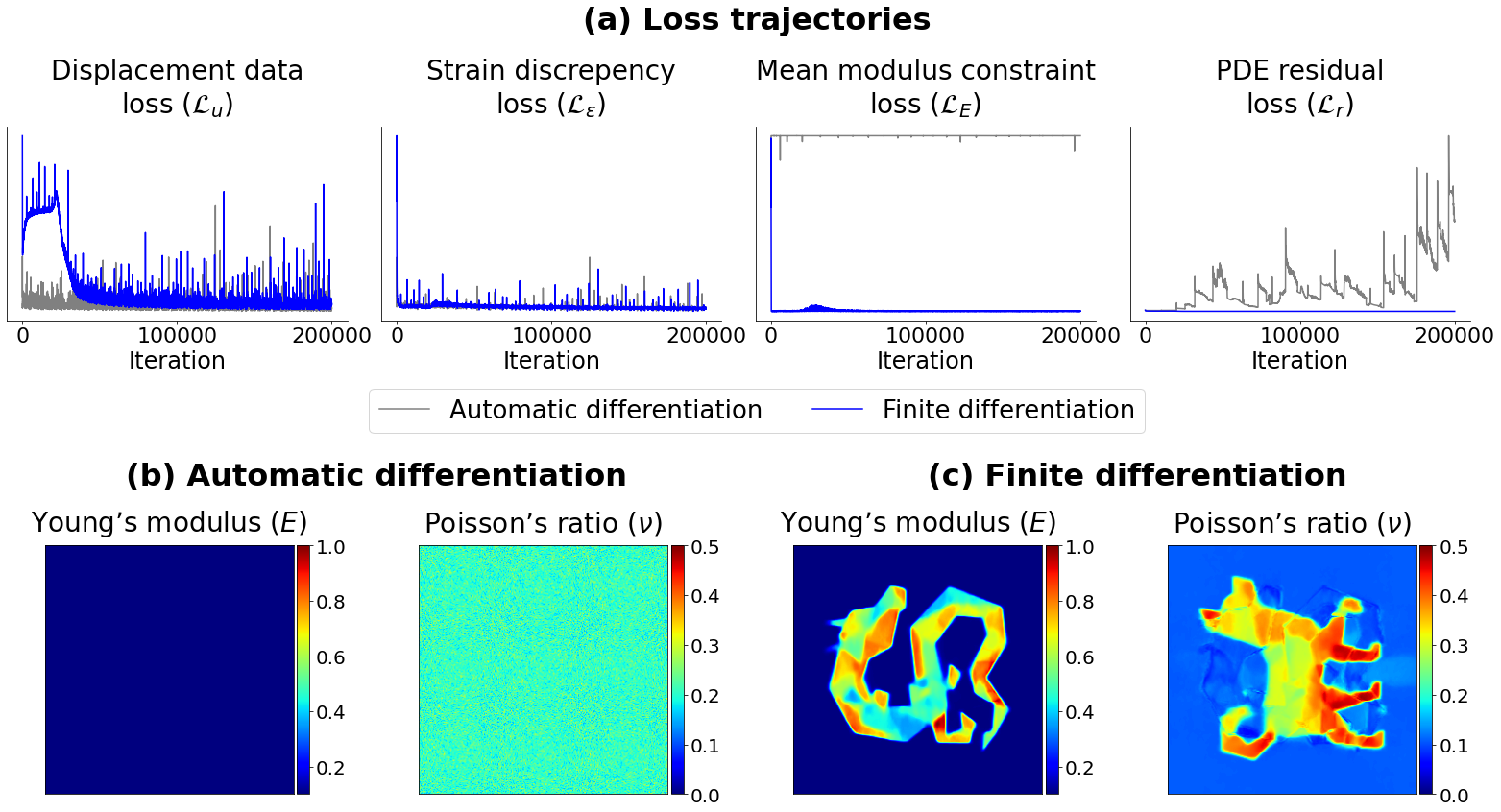}
    \\
    \vspace{0.2 in}
    \caption{ \textbf{Comparison between automatic differentiation and finite differentiation.} Both differentiation methods were applied using the same parameters on the same dataset. The results include (a) Loss trajectories of two different approaches, (b) Elasticity predictions obtained through automatic differentiation, and (c) Elasticity predictions obtained using finite differentiation. In the context of the inverse elasticity problem, automatic differentiation fails to estimate the elasticity parameters.}
    \label{Supple_FDADLoss}

\end{figure}
\newpage
\begin{figure}[h]
    \centering
    \includegraphics[width=1.0\textwidth]{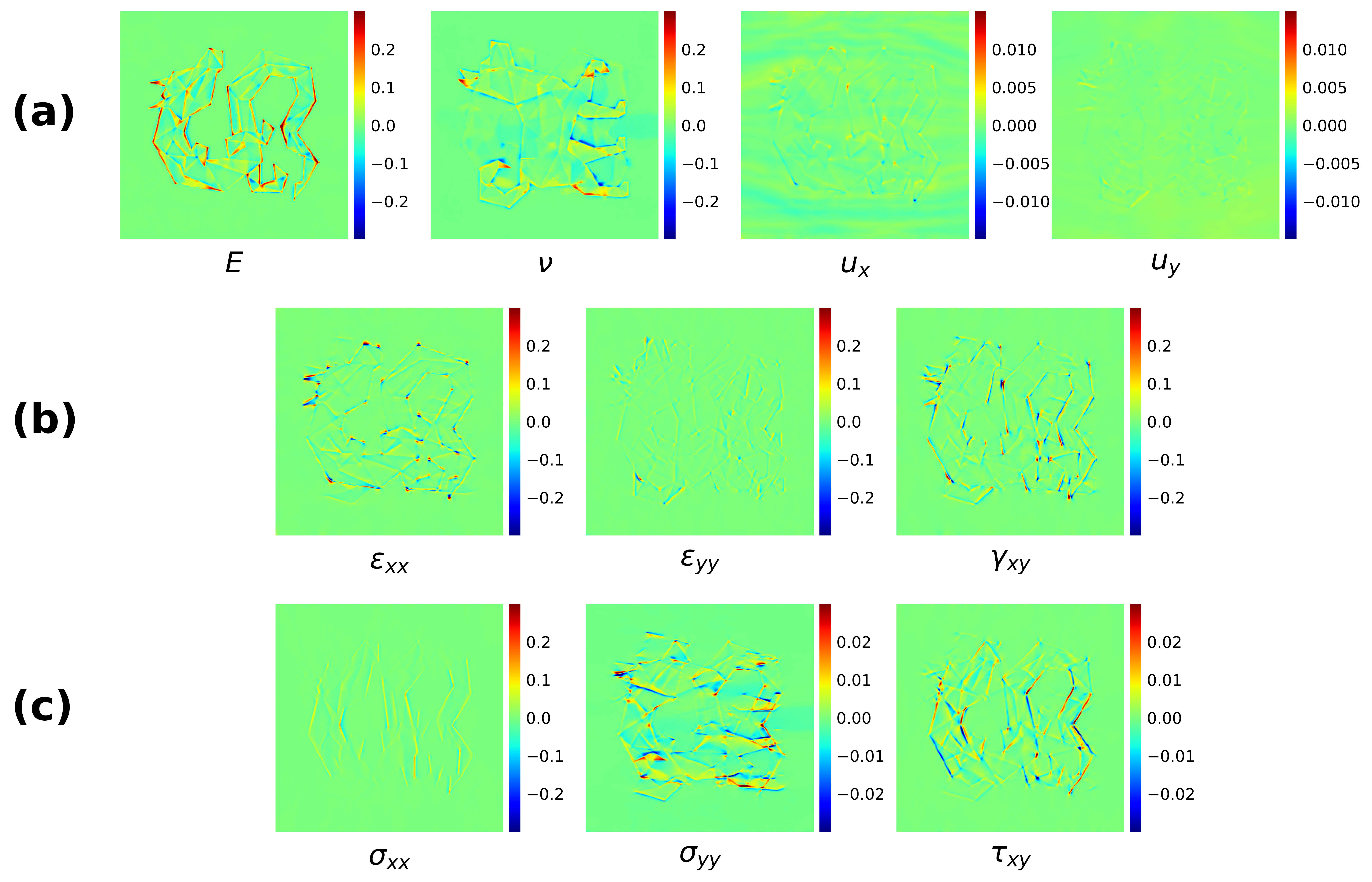}
    \\
    \vspace{0.2 in}
    \caption{ \textbf{The error map of mechanical quantities.} The model is applied with a measured displacement that contains a signal-to-noise ratio of 1000. (a) The error of predicted Young's modulus field (MPa), Poisson's ratio field, and axial displacement field (mm) (b) The error map of predicted strain field ($\%$) (c) The error map of predicted stress field (MPa)}
    \label{Supple_errormap}

\end{figure}
\newpage
\begin{figure}[h]
    \centering
        \begin{subfigure}[t]{0.49\textwidth}
            \centering
            \includegraphics[width=\textwidth]{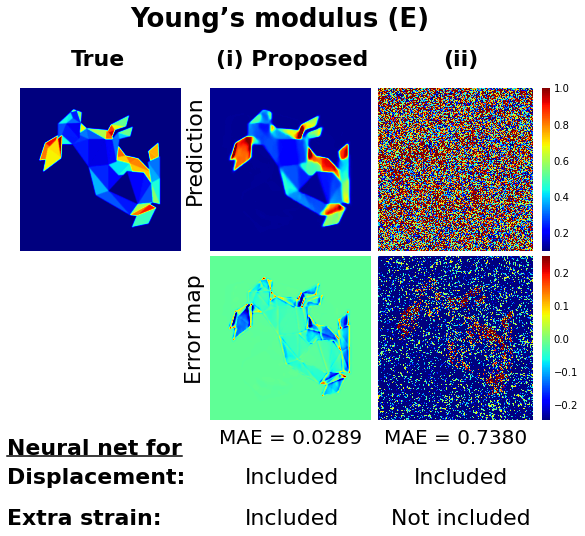} 
            \caption{Estimated Young's modulus across different models.}
            \label{S_RR_1000_E}
        \end{subfigure} 
        \begin{subfigure}[t]{0.49\textwidth}
            \centering
            \includegraphics[width=\textwidth]{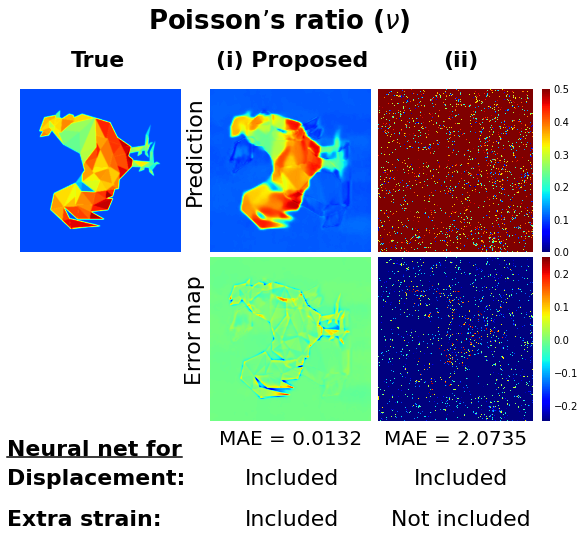} 
            \caption{Estimated Poisson’s ratio across different models.}
            \label{S_RR_1000_v}
        \end{subfigure} 
    \vspace{0.2 in}
    \justifying
    \caption{\textbf{Young's modulus ($E$) and Poisson’s ratio $(\nu)$ prediction and error map of different models.} The models were trained using the same noisy displacement data with a signal-to-noise ratio of 1000, where the true spatial distribution of Young’s modulus has the shape of a rabbit, and that of Poisson’s ratio has the shape of a roaster. (i) IE-PINN (Proposed) includes both displacement and strain networks. (ii) Model only with displacement network (without strain network). In this dataset, Using only displacement networks fails in elasticity estimation.} 

    \label{Supple_RabbitRoaster_SNR1000}
\end{figure}
\newpage

\begin{figure}[h]
    \centering
        \begin{subfigure}[t]{0.49\textwidth}
            \centering
            \includegraphics[width=\textwidth]{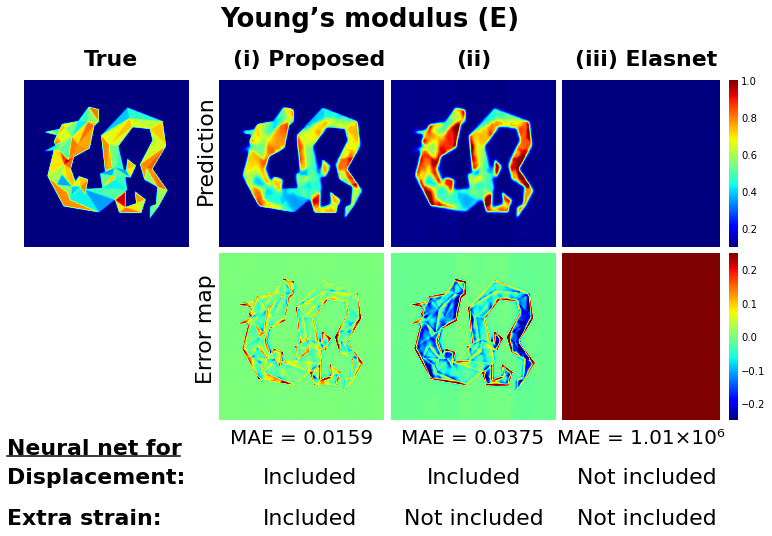} 
            \caption{Estimated Young's modulus across different models.}
            \label{S_DD_500_E}
        \end{subfigure} 
        \begin{subfigure}[t]{0.49\textwidth}
            \centering
            \includegraphics[width=\textwidth]{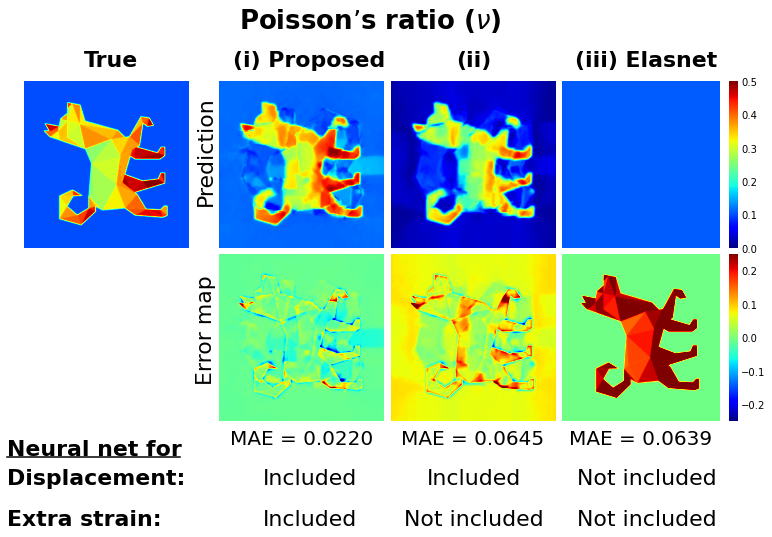} 
            \caption{Estimated Poisson’s ratio across different models.}
            \label{S_DD_500_v}
        \end{subfigure} 
    \vspace{0.2 in}
    \justifying
    \caption{\textbf{Young's modulus ($E$) and Poisson’s ratio $(\nu)$ prediction and error map of different models at a signal-to-noise ratio of 500.} The models were trained using the displacement data with a higher noise level: SNR 500. (i) IE-PINN (Proposed): Model that fits in both displacement and strain. (ii) Model with fitting only displacement data without extra strain fitting and (iii) Elastnet: Model without fitting in both displacement and strain data.\cite{Chen2023} When the noise increases, relying solely on fitting displacement is less effective than additional fitting strain.}
    \label{Supple_DragonDog_SNR500}
\end{figure}
\newpage

\begin{figure}[h]
    \centering
        \begin{subfigure}[t]{0.49\textwidth}
            \centering
            \includegraphics[width=\textwidth]{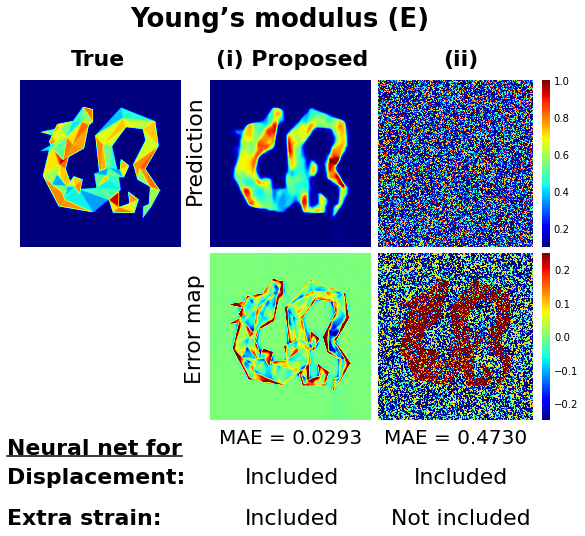} 
            \caption{Estimated Young's modulus across different models.}
            \label{S_DD_100_E}
        \end{subfigure} 
        \begin{subfigure}[t]{0.49\textwidth}
            \centering
            \includegraphics[width=\textwidth]{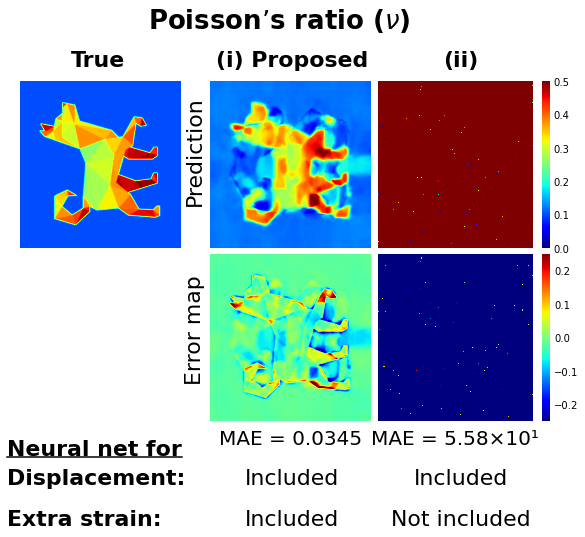} 
            \caption{Estimated Poisson’s ratio across different models.}
            \label{S_DD_100_v}
        \end{subfigure} 
    \vspace{0.2 in}
    \justifying
    \caption{ \textbf{Young's modulus ($E$) and Poisson’s ratio $(\nu)$ prediction and error map of different models at a signal-to-noise ratio of 100.} The models were trained using the displacement data with a higher noise: SNR 100. (i) IE-PINN (Proposed): Model that fits in both displacement and strain. (ii) Model with fitting only displacement data without extra strain fitting, and the strain can be calculated from the displacement-strain relation through finite differentiation. In conditions of high noise, fitting displacement only is not able to resist the high noise condition and leads to worse elasticity prediction.
    \label{Supple_DragonDog_SNR100}
}    
\end{figure}
\newpage

\begin{figure}[h]
    \centering
    \includegraphics[width=0.55\textwidth]{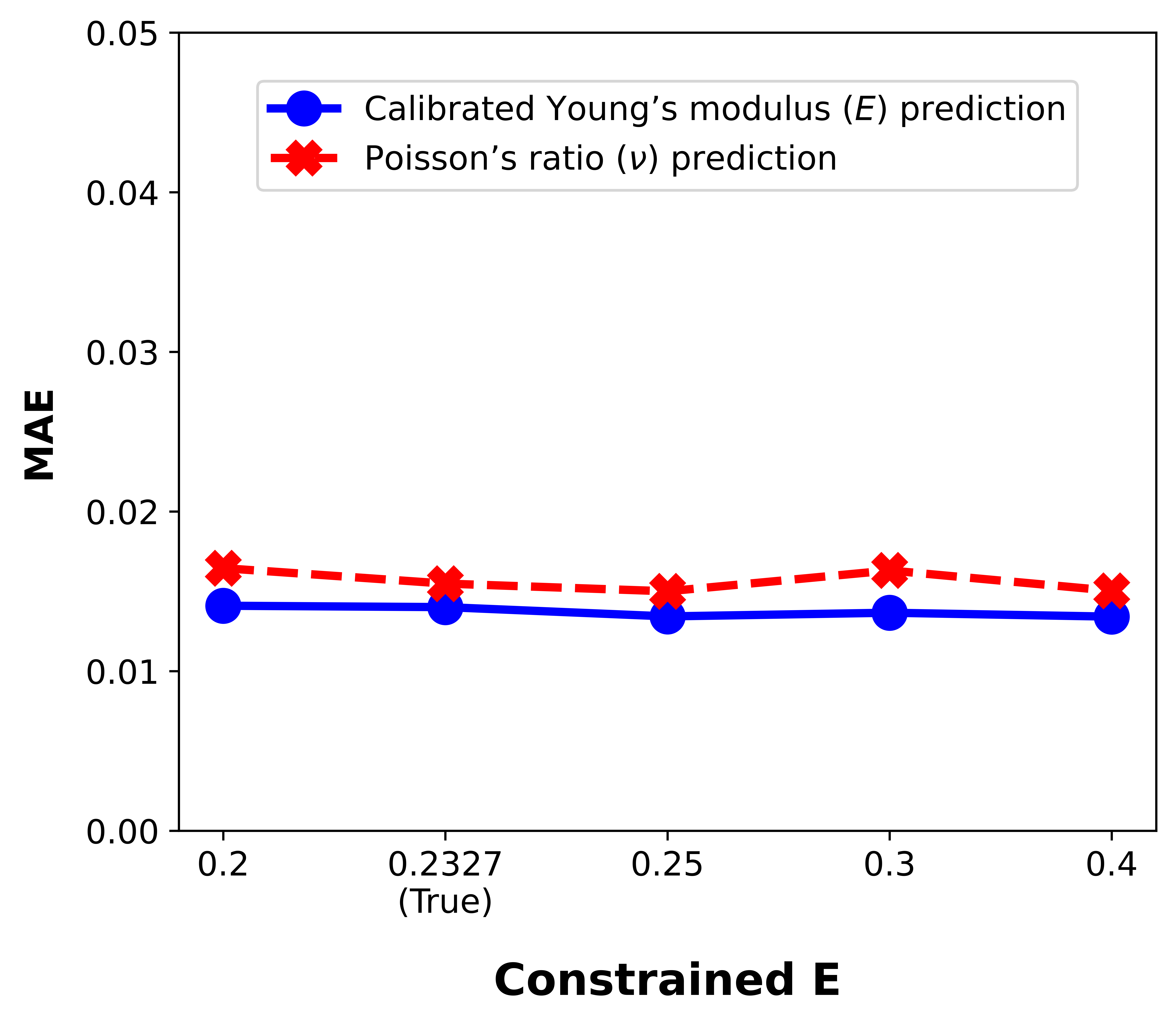} 
    \\
    \vspace{0.2 in}
    \justifying
    \caption{\textbf{Impact of constrained mean Young's modulus to elasticity parameters estimation.} The prediction error (Mean Squared Error: MSE) comparison of  Young's modulus ($E$) and Poisson’s ratio $(\nu)$ predictions of different constrained mean Young's modulus. The model is trained with the Dragon and Dog dataset and is subjected to the measured noisy displacement (SNR1000). Both Young's modulus ($E$) and Poisson’s ratio $(\nu)$ predictions are consistently robust regardless of the different values of mean Young's modulus constraints.}
    \label{Supple_ConstrainedMeanE}
\end{figure}
\newpage

\begin{figure}[h]
    \centering
    \includegraphics[width=0.55\textwidth]{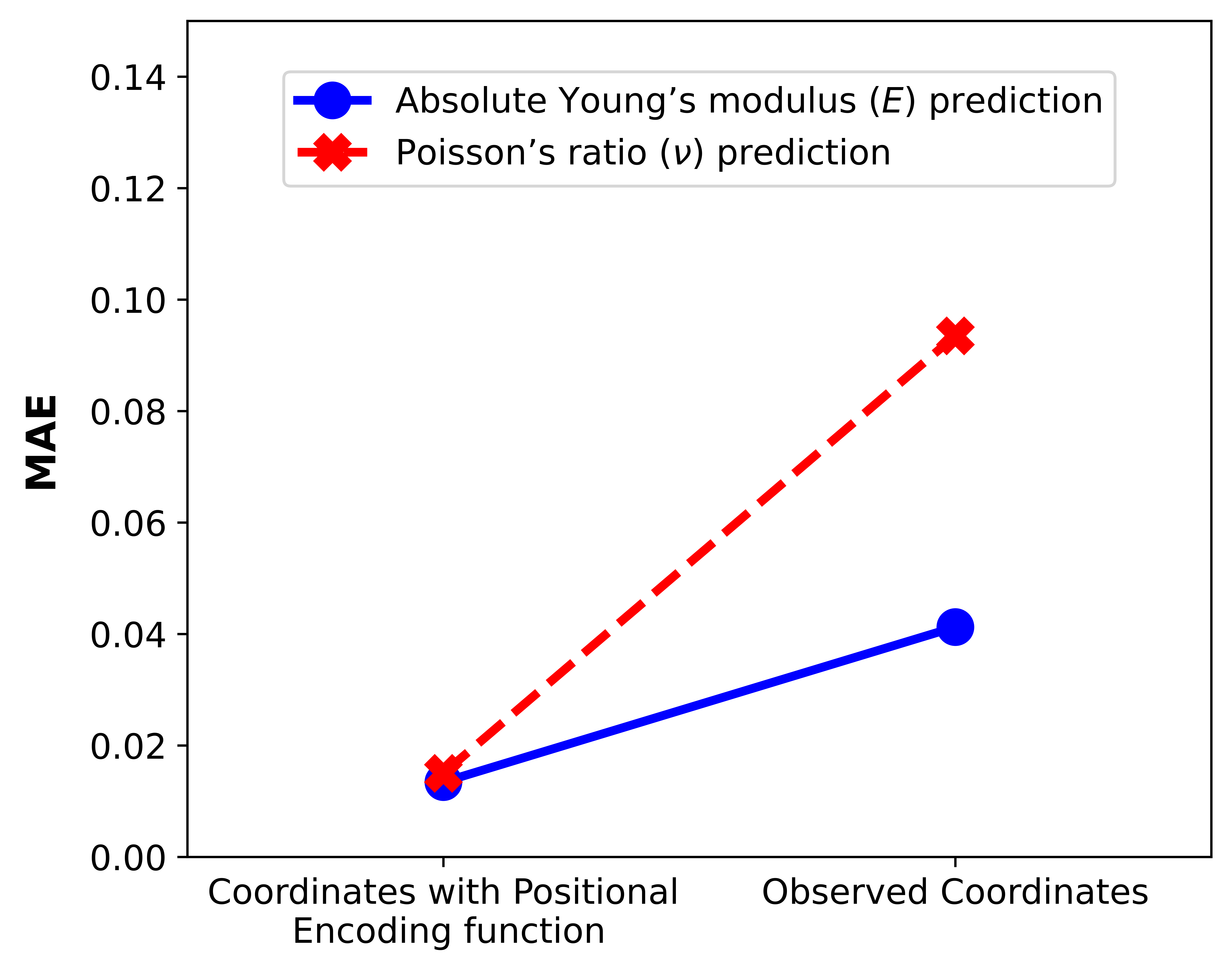} 
    \\
    \vspace{0.2 in}
    \justifying
    \caption{\textbf{Importance of Positional Encoding function.}  The prediction error (Mean Squared Error: MSE) comparison of  Young's modulus ($E$) and Poisson’s ratio $(\nu)$ predictions from coordinates with the Positional Encoding function and observed coordinates in the Dragon and Dog dataset subjected to the measured displacement with a signal-to-noise ratio of 1000. The MRE of Young's modulus ($E$) decreases by half after the Positional Encoding function is integrated into the model. Whereas the MRE of the Poisson’s ratio $(\nu)$ is also significantly improved.}
    \label{Supple_ImportancePositionalEncodingfunction}
\end{figure}
\newpage

\begin{figure}[h]
    \centering
    \includegraphics[width=1.0\textwidth]{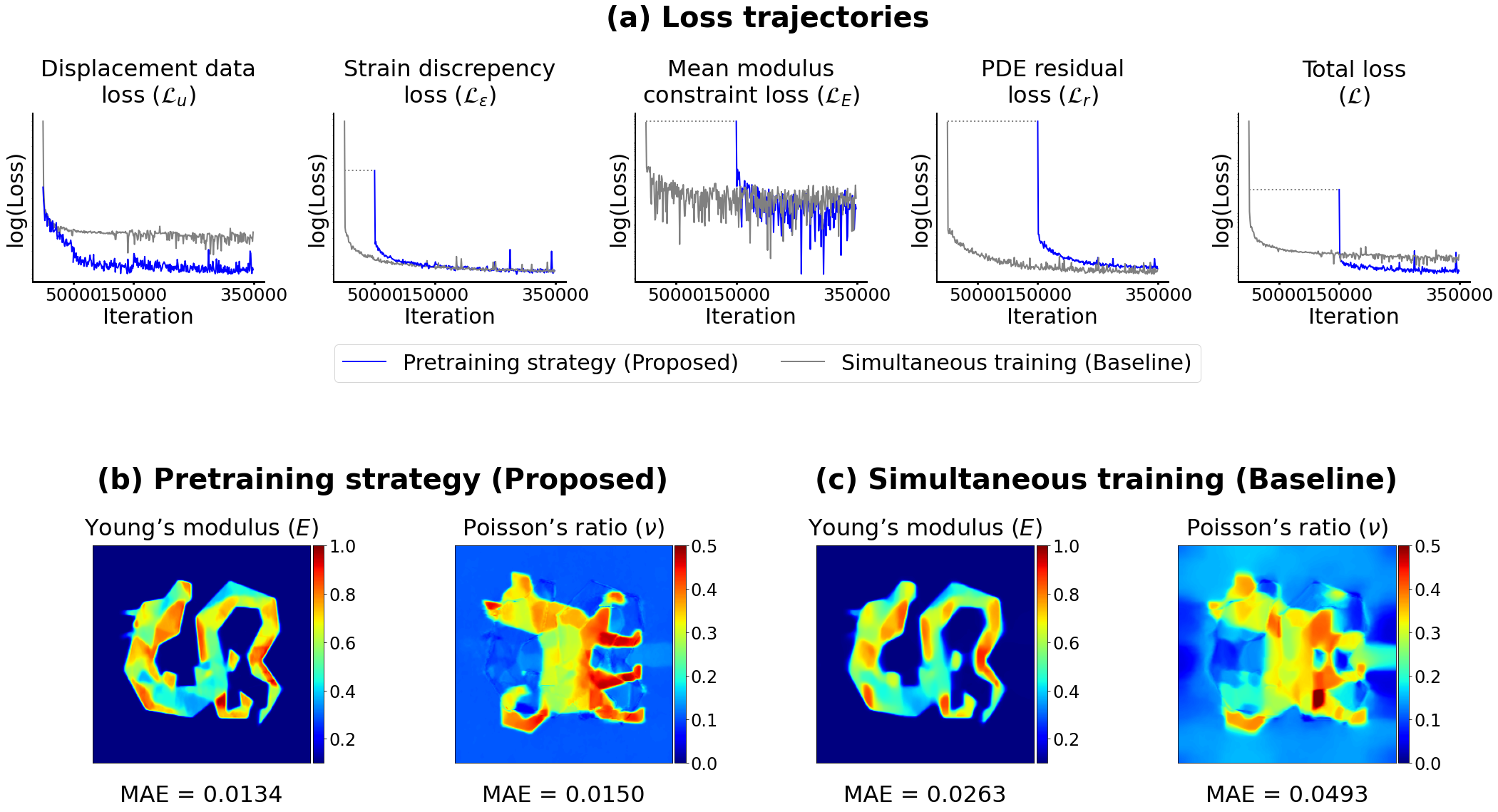} 
    \\
    \vspace{0.2 in}
    \justifying
    \caption{\textbf{Benefits of Pretraining Strategy.} Both training strategies were implemented using the same parameters on the same dataset.The results include: (a) Loss trajectories of two different approaches. The proposed model is trained with pretraining: i) 50,000 iterations for displacement network training, ii) 100,000 iterations with strain network (until 150,000 iterations), and iii) 200,000 iterations with entire networks of IE-PINN. Pretraining effectively reduces the total loss function, especially with significantly lower displacement loss.
    (b) Elasticity predictions obtained from pretraining strategy (Proposed), and (c) Elasticity predictions from simultaneous training of the three-network scheme.The loss trajectories illustrate the trade-offs between the two strategies. While the pretraining strategy can achieve a lower displacement fitting loss, the simultaneous training can achieve a lower PDE residual loss. However, the total loss with the pretraining strategy is significantly better, resulting in more accurate predictions of  Young's modulus ($E$) and Poisson’s ratio $(\nu)$  by half.}
    \label{Supple_Pretraining}
\end{figure}
\newpage
\clearpage
\section*{Supplementary Tables}

\begin{table}[h]
    \centering
        \caption{\textbf{Robustness of accuracy among different constrained mean Young's modulus.} Prediction accuracy of predicted, absolute Young's modulus ($E$) and Poisson’s ratio $(\nu)$ among different constrained mean Young's modulus. The predictions are from the Dragon and Dog dataset subjected to the measured displacement with a signal-to-noise ratio of 1000.}
    \begin{tabular}{|c|c|c|c|}
        \hline
        Constrained $E$  & MAE of Relative $E$ & MAE of Absolute $E$  & MAE of $\nu$ \\ \hline
        0.20 & 0.0340  & 0.0141         & 0.0165       \\
        0.2327 (True) & 0.0152  & 0.0137   & 0.0161    \\
        0.25 & 0.0255  & 0.0134         & 0.0150             \\
        0.30 & 0.0686 & 0.0137        & 0.0163           \\
        0.40 & 0.1669  & 0.0134        & 0.0150            \\
        \hline       
    \end{tabular}
        \\
    \vspace{0.2 in}
    \justifying
    \label{Supple_table_meanE}
\end{table}

\begin{table}[h]
    \centering
    \caption{\textbf{Accuracy improvement with Positional Encoding function.} Prediction accuracy of Young's modulus ($E$) and Poisson’s ratio $(\nu)$ from coordinates with the Positional Encoding function and observed coordinates. The predictions are from the Dragon and Dog dataset subjected to the measured displacement with a signal-to-noise ratio of 1000.}
    \begin{tabular}{|c|c|c|c|}
        \hline
        Input & MAE of $E$ & MAE of $\nu$ \\ \hline
        Positional Encoding function  &  0.0134         & 0.0150           \\
        Actual Coordinates & 0.0412          & 0.0935         \\
        \hline       
    \end{tabular}
        \\
    \vspace{0.2 in}
    \justifying
    \label{Supple_table_positional}
\end{table}
\clearpage
\section*{Supplementary Notes}

\subsection*{Supplementary Note S1: Noisy data generation}
\label{Supple_note_SNR}
\vspace{0.1in}

The datasets for simulation are generated from the published data in the work by Chen and Gu.\cite{Chen2021,Chen2023} The noise is imputed to the raw displacement data as the observed noisy data based on the formulation below.
\begin{align*}
    u' = u + \epsilon  \quad , \quad \epsilon \sim \mathcal{N}(0,\sigma^2) 
\end{align*}
where $u'$ is the observed displacement data imputed with measurement error,$u$ is the actual displacement from the raw data set, and $\epsilon$ is a random noise sampled from the normal distribution with zero mean, and specific standard deviation depending on the signal-to-noise ratio, defined as  
\begin{align*}
    \sigma = \left|{\frac{\bar{u}}{SNR}}\right|
\end{align*}
where $\bar{u}$ is the mean of the displacement measurements and $SNR$ is the signal-to-noise ratio level. This study focused on varying the noise levels from 1000 to 500 and 100. \\

\vspace{0.3in}

\subsection*{Supplementary Note S2: Loading boundary condition} 
   \label{Supple_note_bd}
\vspace{0.1in}

The loading boundary condition refers to the applied force that can cause deformation in an object. This training dataset is generated using a finite element model of an elastic quadrilateral body, where its elasticity is defined by one of 12 zodiac shapes in Young's modulus ($E$), and Poisson’s ratio $(\nu)$.\cite{Chen2021} The top and bottom boundaries of the object are free. In contrast, the left boundary is fixed to restrict horizontal movement (in the x-direction) to zero, allowing only horizontal movement in the right boundary and vertical movement (in the y-direction). The objective of the simulation is to achieve a displacement of 1\% of the body length at the right boundary, which is expected to result in an average normal strain along the x-direction $(\varepsilon_{xx})$ of 1\%. 
Under these conditions, the applied force that defines the loading boundary condition can generally be obtained by the total traction force derived from the stress (calculated from the true boundary strain and elasticity). Since the goal is to produce horizontal displacement at the right boundary, the boundary stress in the x-direction is used to obtain the applied force, as shown in the following equation:
\begin{align*}
    F & \approx \sum_{i=0}^{Y}\sigma_{xx}(x_b,y_i)h
\end{align*}
where $\sigma_{xx}(x_b,y_i)$ is true stress in the x-direction at the boundary $(x_b)$, which is obtained from the elastic constitutive relation based on the predefined elasticity and a true strain of 1\% of the body length, and $h$ is the width of the spacing between two points, which is equal to one according to the dataset.

\end{document}